%% file: main.tex
\documentclass[10pt,twocolumn,letterpaper]{article}

\usepackage{cvpr}              

\usepackage{graphicx}
\usepackage{amsmath}
\usepackage{amsthm}
\usepackage{amssymb}
\usepackage{booktabs}

\usepackage[pagebackref,breaklinks,colorlinks]{hyperref}

\usepackage{url}
\usepackage{multirow}
\usepackage{makecell} 
\usepackage{xcolor}
\usepackage{graphicx}
\usepackage{subcaption}
\usepackage{booktabs} 
\usepackage{bm}
\usepackage{stmaryrd} 
\usepackage{dsfont} 

\usepackage[ruled,vlined]{algorithm2e}

\SetCommentSty{mycommfont}

\def\cvprPaperID{5802} 
\def\confName{CVPR}
\def\confYear{2022}

\begin{document}

\title{StyleGAN-V: A Continuous Video Generator with the Price, Image Quality and Perks of StyleGAN2}

\definecolor{mediumtealblue}{rgb}{0.0, 0.33, 0.71}
\definecolor{darkpastelgreen}{rgb}{0.01, 0.75, 0.24}
\definecolor{azure(colorwheel)}{rgb}{0.0, 0.5, 1.0}

\newcommand{\D}{\mathsf{D}}
\newcommand{\G}{\mathsf{G}}
\newcommand{\F}{\mathsf{F}}
\newcommand{\Ss}{\mathsf{S}}
\newcommand{\ours}{\textcolor{azure(colorwheel)}{(ours)}}
\newcommand{\W}{\mathcal{W}}
\newcommand{\Z}{\mathcal{Z}}
\newcommand{\N}{\mathcal{N}}
\newcommand{\R}{\mathbb{R}}
\newcommand{\zc}{\bm z^\textsf{c}}
\newcommand{\vt}{\bm v_t}
\newcommand{\zm}{\bm z^\textsf{m}}
\newcommand{\Fc}{\mathsf{F}_\text{c}}
\newcommand{\Fm}{\mathsf{F}_\text{m}}
\newcommand{\Nstd}{\mathcal{N}(\bm 0, \bm I)}
\newcommand{\apref}[1]{\ref*{#1}}
\newcommand{\lowerisbetter}{$\downarrow$}

\newcommand{\fix}{\marginpar{FIX}}
\newcommand{\new}{\marginpar{NEW}}
\newcommand{\todo}[1]{\textcolor{red}{TODO: #1}}
\newcommand{\modelname}{StyleGAN-V}
\newcommand{\parallelwork}{(concurrent work)}

\newcommand{\sergey}[1]{{\color{blue}{Sergey:#1}}}
\newcommand{\mohamed}[1]{{\color{orange}{Mohamed:#1}}}
\newcommand{\codeurl}{https://github.com/universome/stylegan-v}
\newcommand{\websiteurl}{https://universome.github.io/stylegan-v}
\newtheorem*{simplestatement*}{A trivial but serviceable statement}

\newcommand{\expect}[2][]{
\ifthenelse{\equal{#1}{}}{
\mathbb{E}\left[#2\right]
}{
\underset{#1}{\mathbb{E}}\left[#2\right]
}}

\author{Ivan Skorokhodov\\
KAUST\\
\and
Sergey Tulyakov\\
Snap Inc.\\
\and
Mohamed Elhoseiny\\
KAUST\\
}
\maketitle

\input{sections/abstract}
\input{sections/introduction}
\input{sections/related-work}
\input{sections/model}
\input{sections/experiments}

\input{sections/conclusion}

{\small
\bibliographystyle{ieee_fullname}
\bibliography{bibliography}
}

\appendix
\clearpage
\input{appendix/limitations}
\input{appendix/training-details}
\input{appendix/evaluation}

\input{appendix/failed-experiments}
\input{appendix/data}
\input{appendix/assumptions}

\input{appendix/additional-samples}
\input{appendix/digan}

\end{document}

%% file: sections/abstract.tex
\begin{abstract}
Videos show continuous events, yet most --- if not all --- video synthesis frameworks treat them discretely in time.
In this work, we think of videos of what they should be --- time-continuous signals, and extend the paradigm of neural representations to build a continuous-time video generator.
For this, we first design continuous motion representations through the lens of positional embeddings.
Then, we explore the question of training on very sparse videos and demonstrate that a good generator can be learned by using as few as 2 frames per clip.
After that, we rethink the traditional image + video discriminators pair and design a holistic discriminator that aggregates temporal information by simply concatenating frames' features.
This decreases the training cost and provides richer learning signal to the generator, making it possible to train \emph{directly} on 1024$^2$ videos for the first time.
We build our model on top of StyleGAN2 and it is just ${\approx}$5\% more expensive to train at the same resolution while achieving almost the same image quality.
Moreover, our latent space features similar properties, enabling spatial manipulations that our method can propagate in time.
We can generate arbitrarily long videos at arbitrary high frame rate, while prior work struggles to generate even 64 frames at a fixed rate. 
Our model is tested on four modern 256$^2$ and one 1024$^2$-resolution video synthesis benchmarks.
In terms of sheer metrics, it performs on average ${\approx}30\%$ better than the closest runner-up.
Project website: \href{\websiteurl}{\websiteurl}.
\end{abstract}

%% file: sections/introduction.tex
\section{Introduction}

\input{figures/long-videos}
\input{figures/manipulation}
\input{figures/compare-to-sg2}

Recent advances in deep learning pushed image generation to the unprecedented photo-realistic quality \cite{StyleGAN2-ADA, BigGAN} and spawned a lot of its industry applications.
Video generation, however, does not enjoy a similar success and struggles to fit complex real-world datasets.
The difficulties are caused not only by the more complex nature of the underlying data distribution, but also due to the computationally intensive video representations employed by modern generators.
They treat videos as discrete sequences of images, which is very demanding for representing long high-resolution videos and induces the use of expensive \texttt{conv3d}-based architectures to model them~\cite{TGAN, MoCoGAN, TGANv2, DVD_GAN}.
\footnote{E.g., DVD-GAN~\cite{DVD_GAN} requires ${\approx}\$30$k to train on $256^2$ resolution~\cite{MoCoGAN-HD}}

In this work, we argue that this design choice is not optimal and propose to treat videos in their natural form: as continuous signals $\bm x(t)$, that map \textit{any} time coordinate $t \in \R_+$ into an image frame $\bm x(t) = \bm x_t \in \R^{3 \times h \times w}$.
Consequently, we develop a GAN-based continuous video synthesis framework by extending the recent paradigm of neural representations \cite{NeRF, SIREN, FourierFeatures} to the video generation domain.

Developing such a framework comes with three challenges.
First, sine/cosine positional embeddings are periodic by design and depend only on the input coordinates.
This does not suit video generation, where temporal information should be aperiodic (otherwise, videos will be cycled) and different for different samples.
Next, since videos are perceived as infinite continuous signals, one needs to develop an appropriate sampling scheme to use them in a practical framework.
Finally, one needs to accordingly redesign the discriminator to work with the new sampling scheme.

To solve the first issue, we develop positional embeddings with time-varying wave parameters which depend on motion information, sampled uniquely for different videos.
This motion information is represented as a sequence of motion codes produced by a \emph{padding-less} \texttt{conv1d}-based model.
We prefer it over the usual LSTM network~\cite{MoCoGAN, MoCoGAN-HD, TGANv2, VideoGLO} to alleviate the RNN's instability when unrolled to large depths and to produce frames non-autoregressively.

Next, we investigate the question of how many samples are needed to learn a meaningful video generator.
We argue that it can be learned from \textit{extremely} sparse videos (as few as 2 frames per clip), and justify it with a simple theoretical exposition (\S\ref{sec:method:sampling}) and practical experiments (see Table~\ref{table:ablations}).

Finally, since our model sees only 2-4 randomly sampled frames per video, it is highly redundant to use expensive \texttt{conv3d}-blocks in the discriminator, which are designed to operate on long sequences of equidistant frames.
That's why we replace it with a \texttt{conv2d}-based model, which aggregates information temporarily via simple concatenation and is conditioned on the time distances between its input frames.
Such redesign improves training efficiency (see Table~\ref{table:main-results}), provides more informative gradient signal to the generator (see Fig~\ref{fig:grads-vis}) and simplifies the overall pipeline (see \S\ref{sec:method:discriminator}), since we no longer need two different discriminators to operate on image and video levels separately, as modern video synthesis models do (e.g., \cite{MoCoGAN, TGANv2, DVD_GAN}).

We build our model, named \modelname, on top of the image-based StyleGAN2~\cite{StyleGAN2}.
It is able to produce arbitrarily long videos at arbitrarily high frame-rate in a non-autoregressive manner and enjoys great training efficiency --- it is only ${\approx}5\%$ costlier than the classical \emph{image-based} StyleGAN2 model~\cite{StyleGAN2}, while having only ${\approx}10\%$ worse \emph{plain} image quality in terms of FID~\cite{FID} (see Fig~\ref{fig:compare-to-sg2}).
This allows us to easily scale it to HQ datasets and we demonstrate that it is \emph{directly} trainable on $1024^2$ resolution.

For empirical evaluation, we use 5 benchmarks: FaceForensics $256^2$~\cite{FaceForensics_dataset}, SkyTimelapse $256^2$~\cite{SkyTimelapse_dataset}, UCF101 $256^2$~\cite{UCF101_dataset}, RainbowJelly $256^2$ (introduced in our work) and MEAD $1024^2$~\cite{MEAD_dataset}.
Apart from our model, we train from scratch 5 different methods and measure their performance using the same evaluation protocol.
Frechet Video Distance (FVD)~\cite{FVD} serves as the main metric for video synthesis, but there is no \emph{complete} official implementation for it (see \S\ref{sec:experiments} and Appx~\apref{ap:fvd}).
This leads to discrepancies in the evaluation procedures used by different works because FVD, similarly to FID~\cite{FID}, is \emph{very} sensitive to data format and sampling strategy~\cite{FID_evaluation}.
That's why we implement, document and release our complete FVD evaluation protocol.
In terms of sheer metrics, our method performs on average ${\approx}30\%$ better than the closest runner-up.

%% file: figures/long-videos.tex
\begin{figure}
    \centering
    \begin{subfigure}[b]{0.49\textwidth}
        \centering
        \includegraphics[width=\textwidth]{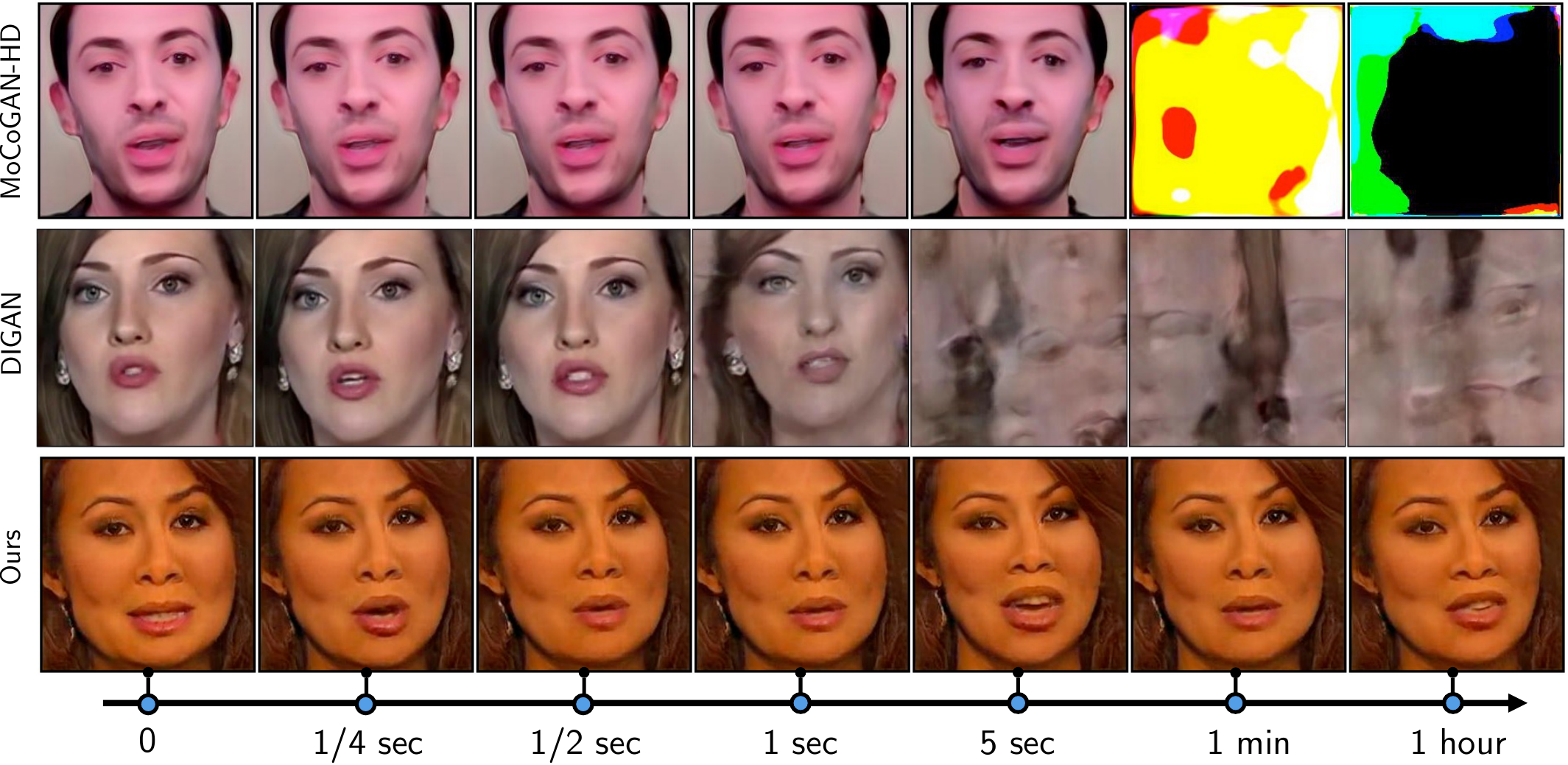}
    \end{subfigure}
    \hfill
    \begin{subfigure}[b]{0.49\textwidth}
        \centering
        \includegraphics[width=\textwidth]{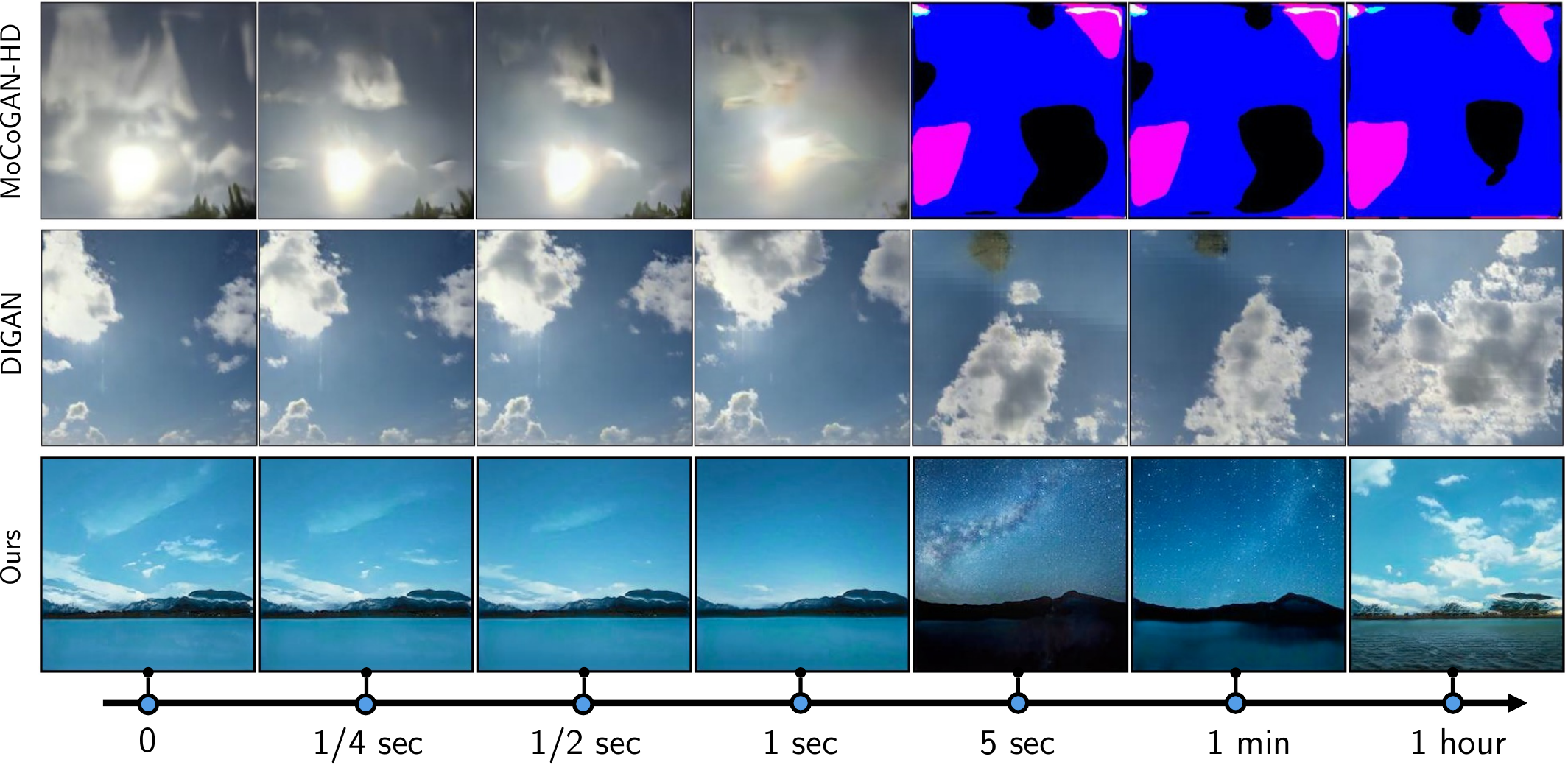}
    \end{subfigure}
    \caption{Examples of 1-hour long videos, generated with different methods. MoCoGAN-HD~\cite{MoCoGAN-HD} fails to generate long videos due to the instability of the underlying LSTM model when unrolled to large lengths. DIGAN~\cite{DIGAN} struggles to generate long videos due to the entanglement of spatial and temporal positional embeddings. StyleGAN-V (our method) generates plausible videos of arbitrary length and frame-rate. Also, unlike DIGAN, it learns temporal patterns not only in terms of motion, but also appearance transformations, like time of day and weather changes.}
    \label{fig:long-videos}
\end{figure}

%% file: figures/manipulation.tex
\begin{figure*}
    \centering
    \includegraphics[width=\linewidth]{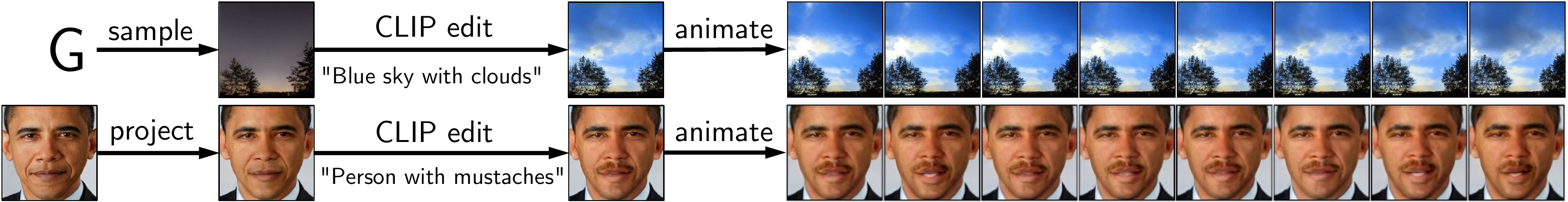}
    \caption{Our model enjoys all the perks of StyleGAN2~\cite{StyleGAN2}, including the ability of semantic manipulation. In this example, we edited a generated frame (top row) or projected off-the-shelf image (bottom row) with CLIP and animated it with our model. To the best of our knowledge, our work is the first one which demonstrates such capabilities for video generators.}
    \label{fig:latent-manipulation}
\end{figure*}

%% file: figures/compare-to-sg2.tex
\begin{figure}
\centering
\begin{subfigure}[b]{0.49\linewidth}
    \centering
    \includegraphics[width=\textwidth]{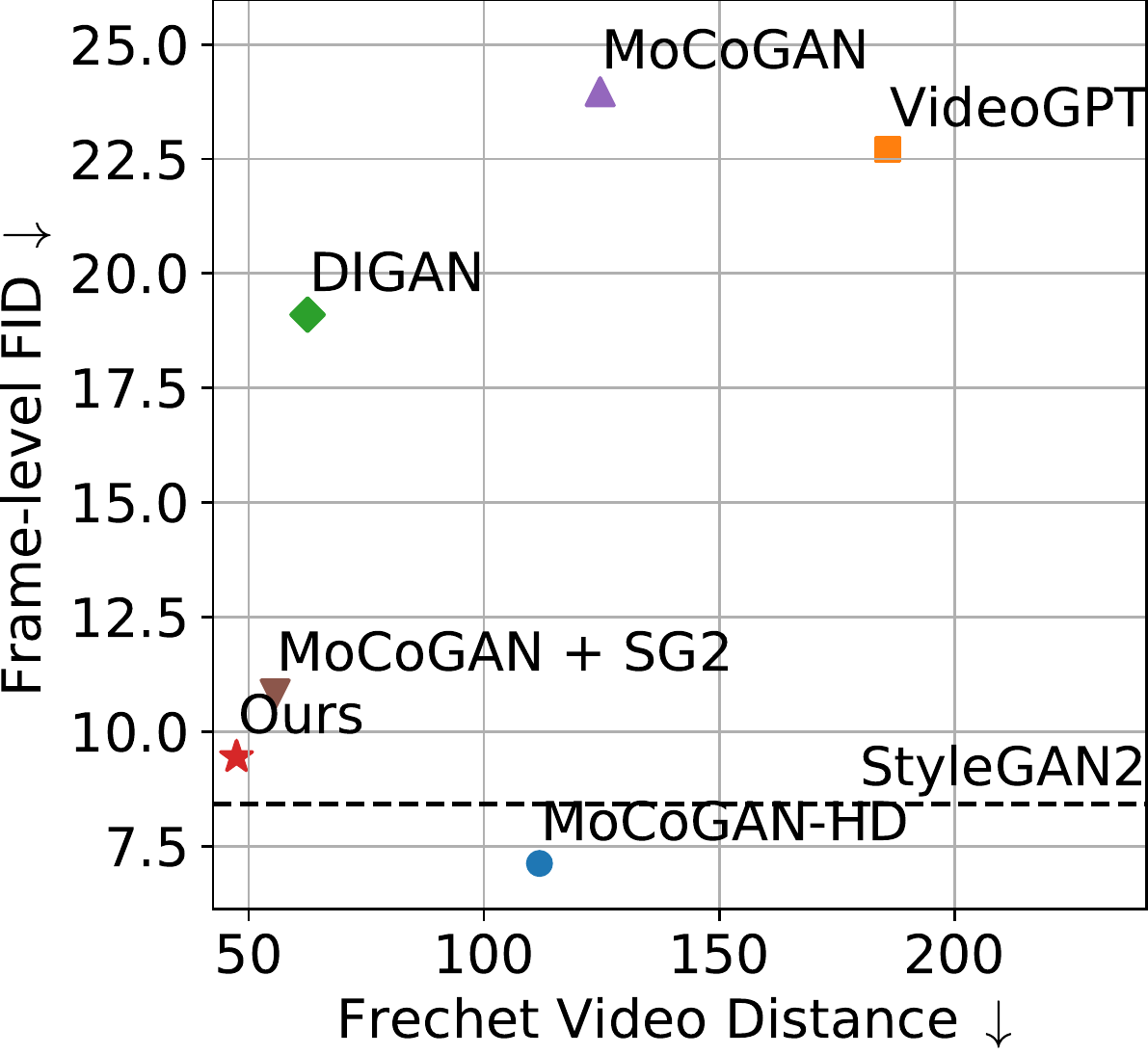}
\end{subfigure}
\hfill
\begin{subfigure}[b]{0.478\linewidth}
    \centering
    \includegraphics[width=\textwidth]{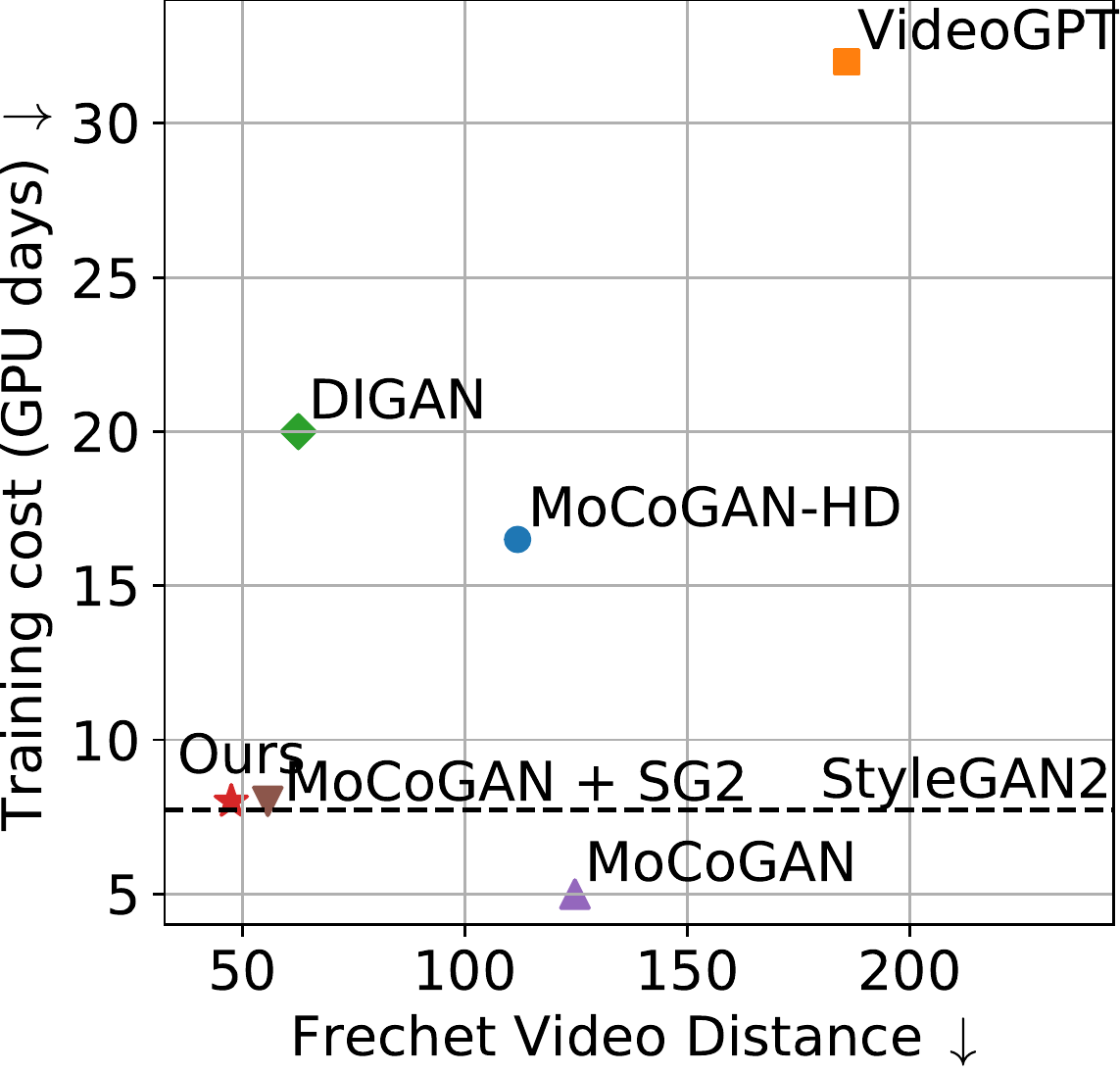}
\end{subfigure}
\caption{FID scores and training cost by FVD$_{16}$ for modern video generators on FaceForensics $256^2$~\cite{FaceForensics_dataset}. Our method (denoted by {\normalsize\textcolor{red}{$\star$}}) shows that video generators can be as efficient and as good in terms of image quality as traditional image based generators (like, StyleGAN2~\cite{StyleGAN2}, denoted with the dashed line).}
\label{fig:compare-to-sg2}
\end{figure}

%% file: sections/related-work.tex
\section{Related work}


\textbf{Video synthesis}.
Early works on video synthesis mainly focused on \textit{video prediction} \cite{SiftFlow, UnsupervisedVisualPrediction}, i.e. generating future frames given a sequence of the previously seen ones.
Early approaches for this problem typically employed recurrent convolutional models trained with reconstruction objective \cite{video_language_modeling, Robot_pushing_dataset, LSTMs_video_representations}, but later adversarial losses were introduced to improve the synthesis quality \cite{MultiScaleVideoPrediction, Video2Video, GeneratingVideosWithSceneDynamics}.
Some recent works explore autoregressive video prediction with recurrent or attention-based models (e.g., \cite{VideoTransformer, LVT, VideoGPT, PredictingVideoWithVQVAE, VideoPixelNetworks}).
Another close line of research is \textit{video interpolation}, i.e. increasing the frame rate of a given video (e.g., \cite{Video_interpolation_ASC, Video_interpolation_DAIN, Video_interpolation_SuperSloMo}).
In our work, we study \textit{video generation}, which is a more challenging problem than video prediction since it seeks to synthesize videos from scratch, i.e. without using the expressive conditioning on previous frames.
Classical methods in this direction are typically based on GANs \cite{GANs}.
MoCoGAN~\cite{MoCoGAN} and TGAN~\cite{TGAN} decompose generator's input noise into a content code and motion codes, which became a standard strategy for many subsequent works (e.g., \cite{MoCoGAN-HD, TGANv2, VideoGLO, TemporalShiftGAN}).
Several approaches consider video generation from a single clip (e.g., \cite{HP-VAE-GAN, VGPNN, SinGAN-GIF}).

Some recent works also consider high-resolution video synthesis~\cite{MoCoGAN-HD, StyleVideoGAN}, but only with training in the latent space of a pretrained image generator.
\modelname\ is trained on \textit{extremely} sparse videos.
This makes it related to \cite{TGANv2, Hierarchical_video_generation, Inmodegan}, which use a pyramid of discriminators operating on different temporal resolutions (with a subsampling factor of up to $\times 8$).
Our model builds on the time continuity, which in the context of video synthesis was also explored by \cite{VidODE}.

To the best of our knowledge, all modern video synthesis approaches utilize expensive \texttt{conv3d} blocks either in their decoder and/or encoder components (e.g., \cite{MoCoGAN, MoCoGAN-HD, TGANv2, TemporalShiftGAN, DVD_GAN, LDVDGAN, ProVGAN}).
Often, GAN-based approaches utilize two discriminators, operating on image and video levels independently, where the video discriminator operates at a low resolution to save computation (e.g., \cite{MoCoGAN, G3AN, MoCoGAN-HD, DVD_GAN}).
In our work, we aggregate the temporal information via a simple concatenation of feature vectors extracted from the frames and this strategy suffices to build a state-of-the-art video generator.

\textbf{Neural Representations}.
Neural representations is a recent paradigm that uses neural networks to represent continuous signals, such as images, videos, audios, 3D objects and scenes (e.g., \cite{NeRF, SIREN, FourierFeatures, SRNs, TemplateImplicitFunction}).
It is mostly popular for 3D reconstruction and geometry processing tasks (e.g., \cite{DeepSDF, DeepMeta, OccupancyNetworks, ConvolutionalOccupancyNetworks, DVR}), including video-based reconstruction~\cite{Nerfies, SpaceTimeNeIF, D-nerf, NSFF}.
Several recent projects explored the task of building generative models over such representations to synthesize images (e.g., \cite{INR-GAN, CIPS, ALIS}), 3D objects (e.g., \cite{GRAF, piGAN, NeRF-VAE}) or multi-modal signals (e.g., \cite{INRs_distribution, GEM}), and our work extends this line of research to video generation.


\textbf{Concurrent works}.
The development of neural representations-based approaches moves extremely fast and there are two concurrent works which propose ideas similar to our ones.
DIGAN\cite{DIGAN} is a concurrent project that explores the same direction of using neural-based representations for continuous video synthesis and shares a \emph{lot} of ideas with our work.
The authors also consider a continuous-time generator, trained by a discriminator without \texttt{conv3d} layers.
The core difference with our work is that they use a different parametrization of motions and use a dual discriminator $\D$: one operates on $(\bm x_1, \bm x_2, \Delta t)$ and the second one on individual images.
We enumerate the differences and similarities in Appx~\apref{ap:digan}.
NeRV~\cite{NeRV} uses convolutional neural representations of videos for compression and denoising tasks.
GEM~\cite{GEM} utilizes generative latent optimization~\cite{GLO} to build a multi-modal generative model.

%% file: sections/model.tex
\section{Model}\label{sec:method}
\input{figures/grads-vis}


Our model is based on the paradigm of \textit{neural representations} \cite{NeRF, SIREN, FourierFeatures}, i.e. representing signals as neural networks.
We treat each video as a function $\bm x_t = \bm x(t)$ which is continuous in time $t \in \R_+$.
In this manner, the training dataset $\mathcal{D}$ is a set of subsampled signals
$\mathcal{D} = \{ \bm x^{(i)} \}_{i=1}^N = \{ (\bm x_{t_0}^{(i)}, ..., \bm x^{(i)}_{t_{\ell_i}}) \}_{i=1}^N$, where $N$ denotes the total number of videos, $t_j$ denotes the time position of the $j$-th frame and $\ell_i$ is the amount of frames in the $i$-th video.\footnote{To simplify the notation, we assume that all videos have the same frame-rate and that all the videos were sampled starting at $t = t_0$.}
Note that each video might have a different length $\ell_i$ and in practice these lengths vary a lot (see Appx~\apref{ap:data} for datasets statistics).
Our goal is to train a generative model over video signals, having only their subsampled versions.
To achieve this, we develop the following framework.



We build the model on top of StyleGAN2 ~\cite{StyleGAN2-ADA} and redesign its generator and discriminator networks for video synthesis with minimal modifications.
Our generator is conceptually similar to MoCoGAN~\cite{MoCoGAN}, i.e., we separate latent information into content code $\zc$ and motion trajectory $\vt = \bm v(t)$.
In contrast to MoCoGAN, our motion codes $\vt$ are continuous in time $t \in \R_+$ and we describe their design in \S\ref{sec:method:generator}.
The \textit{only} modification we do on top of StyleGAN2's generator is the concatenation of our continuous motion codes $\bm v_t$ to its constant input tensor.
The discriminator model $\D$ takes $k$ frames $\bm x_{t_1}, ..., \bm x_{t_k}$ of a sparsely sampled video, independently extracts features $\bm h_{t_1}, ..., \bm h_{t_k}$ from them, concatenates those features together channel-wise into a global video descriptor $\bm h$ and predicts the real/fake class from it.
We condition $\D$ on the time distances $\delta^x_i = t_{i+1} - t_i$ between frames to make it easier for it to operate on different frame rates.


\input{figures/generator}

\subsection{Generator structure}\label{sec:method:generator}

\textbf{Overview}.
Generator consists of three components: content mapping network $\Fc$, motion mapping network $\Fm$ and synthesis network $\Ss$.
$\Fc$ and $\Ss$ are borrowed from StyleGAN2 and we only modify $\Ss$ by tiling and concatenating motion codes $\vt$ to its constant input tensor.

A video is generated the following way.
First, we sample the content noise $\zc \sim \Nstd$ and, following StyleGAN2, transform it into latent code $\bm w = \Fc(\zc) \in \R^{512}$.
It is shared for all timesteps $t \in \R_+$ of a video.
Then, to generate a frame $\bm x_t$ in the specified time location $t$, we first compute its motion code $\bm v_t$, which is done in three steps.
First, we sample a discrete sequence of equidistant trajectory noise $\zm_{t_0}, ..., \zm_{t_n} \sim \Nstd$ (we assume $t_0 = 0$ everywhere), positioned at distance $\delta^z = t_{i+1} - t_i$ from one another.
The number of tokens $n$ is determined by the condition $t < t_n$, i.e. it should be long enough to cover the desired timestep $t$.\footnote{In practice, since $\Fm$ uses padding-less convolutions, this sequence is slightly larger. We elaborate on this in Appx~\ref{ap:training-details}.}
Then, we process it with \texttt{conv1d}-based motion mapping network $\Fm$ with a large kernel size into the sequence $\bm u_{t_0}, ..., \bm u_{t_n}$.
After that, we take a pair of tokens $\bm u_\ell, \bm u_r$ which $t$ lies between (i.e. $\ell = t_i$ for some $i \in \{0, 1, ..., n\}$ and $r = t_{i + 1}$) and compute an acyclic positional embedding $\bm v_t$ from them, described next.
This positional embedding serves as the motion code for our generator.
In fact, we do not need to sample all the motion noise vectors $\zm_{t_0}, ..., \zm_{t_n}$ to produce $\bm v_t$, but only those ones which $\bm v_t$ depends on.
In this way, our generator can produce frames non-autoregressively.

\textbf{Acyclic positional encoding}.
Traditional positional embeddings~\cite{SIREN, FourierFeatures} are cyclic by default.
This does not create problems in traditional applications (like image or scene representations) because utilized spatial domain there never exceeds the period length~\cite{NeRF, INR-GAN}.
But for video generation, cyclicity is not desirable, because it makes a video getting looped at some point.
To solve this issue, we develop \textit{acyclic} positional encoding.

A sine-based positional embedding vector $\bm p \in \R^d$ can be expressed in the following form:
\begin{equation}
\bm p(\bm\alpha, \bm\omega, \bm\rho, t) = \bm\alpha \odot \sin(\bm\omega \cdot t + \bm\rho),
\end{equation}
where $\odot$ denotes element-wise vector multiplication, $\bm \alpha, \bm\omega, \bm\rho \in \R^d$ are amplitudes, periods and phases of the corresponding waves, and the sine function is applied element-wise.
By default, these embeddings are periodic and always the same for any input \cite{SIREN, FourierFeatures, NeRF}, which is not desirable for video synthesis, where natural videos contain different motions and are typically aperiodic.
To solve this issue, we compute the wave parameters from motion noise $\zm_{t_0}, ..., \zm_{t_n}, ...$ the following way.
First, ``raw'' motion codes $\tilde{\bm v}_t$ are computed using wave parameters $\bm\alpha_\ell, \bm\omega_\ell, \bm\rho_\ell$ predicted from $\bm u_\ell$:
\begin{equation}
\tilde{\bm v}_t = \bm \alpha_\ell \odot \sin(\bm\omega_\ell \cdot t + \bm\rho_\ell),
\end{equation}
where
\begin{equation}
\bm \alpha_\ell = W_\alpha \bm u_\ell, \quad
\bm \omega_\ell = W_\omega \bm u_\ell, \quad
\bm \rho_\ell = W_\rho \bm u_\ell,
\end{equation}
and $W_\alpha, W_\omega, W_\rho \in \R^{d \times d}$ are learnable weight matrices.
Using $\tilde{\bm v}_t$ directly as motion codes does not lead to good results since it contains discontinuities (see Fig~\apref{fig:time-embs:ours-raw}).
That's why we ``stitch'' their start and end values via:
\begin{equation}\label{eq:alignment}
\bm v_t = \tilde{\bm v}_t - \text{lerp}(\tilde{\bm v}_\ell, \tilde{\bm v}_r, t) + \text{lerp}(W_a \bm u_\ell, W_a \bm u_r, t),
\end{equation}
where $W_a \in \R^{d \times d}$ is a learnable weight matrix and \texttt{lerp}$(\bm x, \bm y, t)$ is the element-wise linear interpolation between $\bm x$ and $\bm y$ using the time position $t$.
The first subtraction in Eq~\eqref{eq:alignment} alters the positional embeddings to make them converge to zero values at locations $\{t_0, t_1, ..., t_n, ...\}$.
This limits the expressive power of the positional embeddings and that's why we add the ``alignment'' vectors $\bm a = W_a \bm u$ to restore it.
See Fig~\apref{fig:time-embs:ours-no-aligners} in Appx~\apref{ap:training-details} for the visualization.

In practice, we found it useful to compute periods as:
\begin{equation}
\bm\omega_t = (\text{tanh}(W_\omega \bm u_t) + \mathds{1}) \odot \bm\sigma,
\end{equation}
where $\mathds{1}$ is a vector of ones and $\bm\sigma$ are linearly-spaced scaling coefficients.
See Appx~\apref{ap:training-details} and the source code for details.


One could try using continuous codes $\bm u_t = \texttt{lerp}(\bm u_\ell, \bm u_r, t)$ directly as motion codes instead of $\bm v_t$.
This also eliminates cyclicity (in theory), but leads to poor results in practice: if the distance $\delta^z$ is small, then the motion trajectory will contain unnatural sharp transitions; and when $\delta^z$ is increased, $\G$ loses its ability to properly model high-frequency motions (like blinking) since the codes change too slowly.
We empirically validate this in Tab~\ref{table:ablations} (also see samples on the project webpage).

\subsection{Discriminator structure}\label{sec:method:discriminator}
Modern video generators typically utilize two separate discriminators which operate on image and video levels separately \cite{DVD_GAN, MoCoGAN, MoCoGAN-HD}.
But since we train on extremely sparse videos and aim to have a computationally efficient model, we propose to use a \textit{holistic} discriminator $\D(\bm x_{t_1}, ..., \bm x_{t_k})$, which is conditioned on the time distances between frames $\delta^x_i = t_{i+1} - t_i$.
It consists of two parts: 1) feature extractor backbone $\D_b$, which independently embeds an image frame $\bm x_t$ into a 3D feature vector $\bm h_{t_i} \in \R^{512 \times 16 \times 16}$; and the convolutional head $\D_h$, which takes the concatenation of all the features $\bm h = \texttt{concat}[\bm h_{t_1}, ..., \bm h_{t_k}] \in \R^{512k \times 16 \times 16}$ and outputs the real/fake logit $y \in \R$.

We input the time distances information $\delta^x_{1}, ..., \delta^x_{k-1}$ between $k$ frames $\bm x_{t_1}, ..., \bm x_{t_{k}}$ into $\D$ the following way.
First, we encode them with positional encoding, preprocess with a 2-layer MLP into $\bm p(\delta^x_{1}), ..., \bm p(\delta^x_{k-1}) \in R^d$ and concatenate into a single vector $\bm p_\delta \in \R^{(k-1) \cdot d}$.
After that, we use the projection discriminator strategy~\cite{ProjectionDiscriminator} and compute the output logit as a simple dot product between $\bm p_\delta$ and the corresponding video feature vector.
The overall architecture is visualized in Fig~\ref{fig:discriminator}.

\input{figures/discriminator}

Such a design is \textit{greatly} more efficient than using both image and video discriminators
and provides a more informative learning signal to the generator (see Fig~\ref{fig:grads-vis}).

\input{figures/samples}

\subsection{Implicit assumptions of sparse training}\label{sec:method:sampling}


Consider the problem of learning a probability distribution $p(\bm x) = p(x_1, ..., x_n)$ and consider that we utilize sparse training, i.e. select $k$ coordinates of vector $\bm x$ randomly on each iteration of the optimization process.
Then the optimization objective is equivalent to learning all possible marginal distributions $p(x_{i_1}, ...., x_{x_k})$ instead of learning joint $p(\bm x)$.
When does learning marginals allow to obtain the full joint distribution at the end?
The following simple statement adds some clarity to this question.
\begin{simplestatement*}
Let's denote by $\mathcal{J}_{<i}^k$ a collection of sets $J_i$ of up to $k$ indices $j$ s.t. $\forall J_i \in \mathcal{J}_{<i}^k$ we have $j < i$ for all $j \in J_i$.
In other words, $J_i$ is a set of up to $k$ indices $j \in [1, i)$.
Then, $p(\bm x)$ can be represented as a product of $n$ marginals $p(x_i, \bm x_{J_i})$ for $i \in [1, n]$ if and only if $\forall i$ there exists $J_i \in \mathcal{J}_{<i}^{k-1}$ s.t. $p(x_i | \bm x_{<i}) \equiv p(x_i | \bm x_{J_i})$.
\end{simplestatement*}

The above statement is primitive (see the proof in Appx~\apref{ap:assumptions}) but can provide useful practical intuition.
For video synthesis, it implies that one can learn a video generator by using only $k$ frames per video only if for any frame $\bm x_i$, there exists \emph{at most} $k - 1$ previous frames sufficient to properly predict it (see Appx~\apref{ap:assumptions}).
And we argue that very few frames suffice to make such a prediction for the modern video synthesis benchmarks.
For example, in SkyTimelapse~\cite{SkyTimelapse_dataset}, the motions are typically unidirectional and thus easily predictable from only 2 previous frames, which corresponds to training with $k=3$ frames per video.

We treat videos as infinite continuous signals, but in practice one has to set a limit on the maximum time location $T$ which can be seen during training.
To the best of our knowledge, previous methods use at most $T=64$ \cite{TGANv2, Hierarchical_video_generation}, but in our case we easily train the model with $T=1024$ since our generator is non-autoregressive and our discriminator uses only the relative temporal information.
We set the maximum distance between $t_1$ and $t_k$ to 32 to cover short and medium-term movements: otherwise, we observed unstable training and abrupt motions.
To sample frames, we first sample the distance $(t_k - t_1) \sim U[k-1, 32]$ between them, and then sample the offset $t_1 \sim U[0, T - t_k]$.
After that, frames locations $t_i$ for $i \in \{2, ..., k-1\}$ are selected at random without repetitions.

%% file: figures/grads-vis.tex
\begin{figure}
    \centering
    \includegraphics[width=\linewidth]{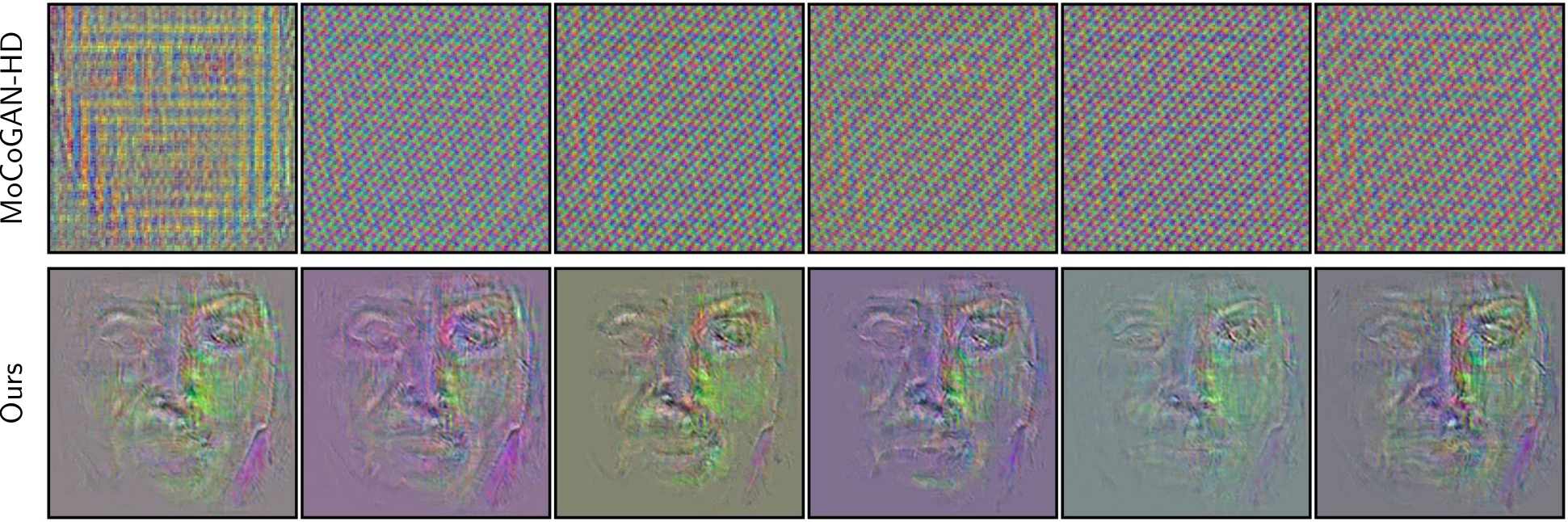}
    \caption{Visualizing the gradient signal to $\G$ at ${\approx}50\%$ of training from \texttt{conv3d}-based discriminator of MoCoGAN-HD (upper row) and our one (lower row) at $t = 0, 2, 4, 6, 8, 12$ timesteps.}
    \label{fig:grads-vis}
\end{figure}

%% file: figures/generator.tex
\begin{figure}
    \centering
    \includegraphics[width=1.0\linewidth]{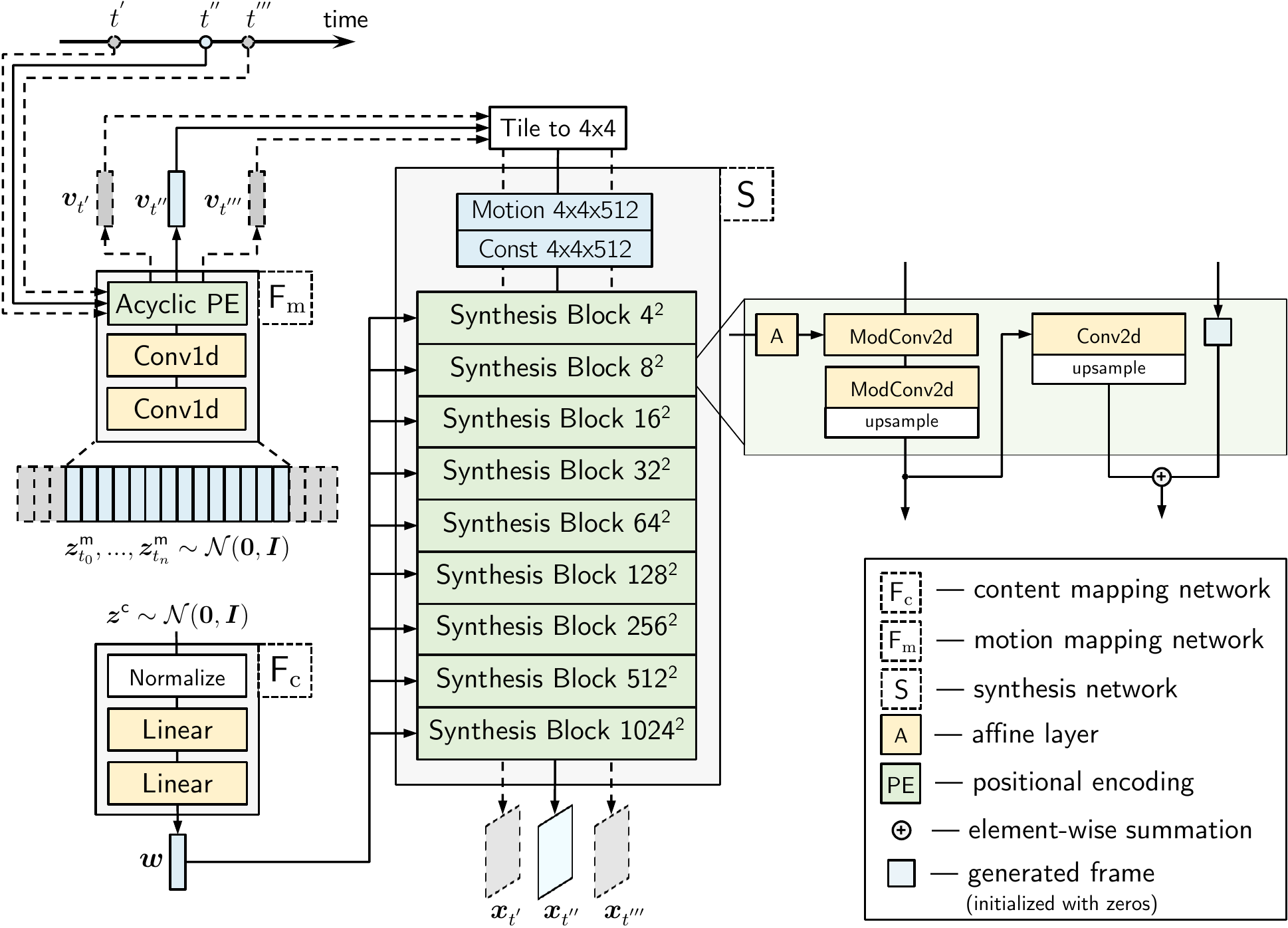}
    \caption{Generator architecture: the only change we do on top of StyleGAN2 generator's synthesis network $\Ss$ is the concatenation of our motion codes to the constant input tensor. $\Ss$ produces frames $\bm x_t$ non-autoregressively using the content code $\bm w$ and motion code $\bm v_t$.}
    \label{fig:motion-mn}
\end{figure}

%% file: figures/discriminator.tex
\begin{figure}
    \centering
    \includegraphics[width=0.9\linewidth]{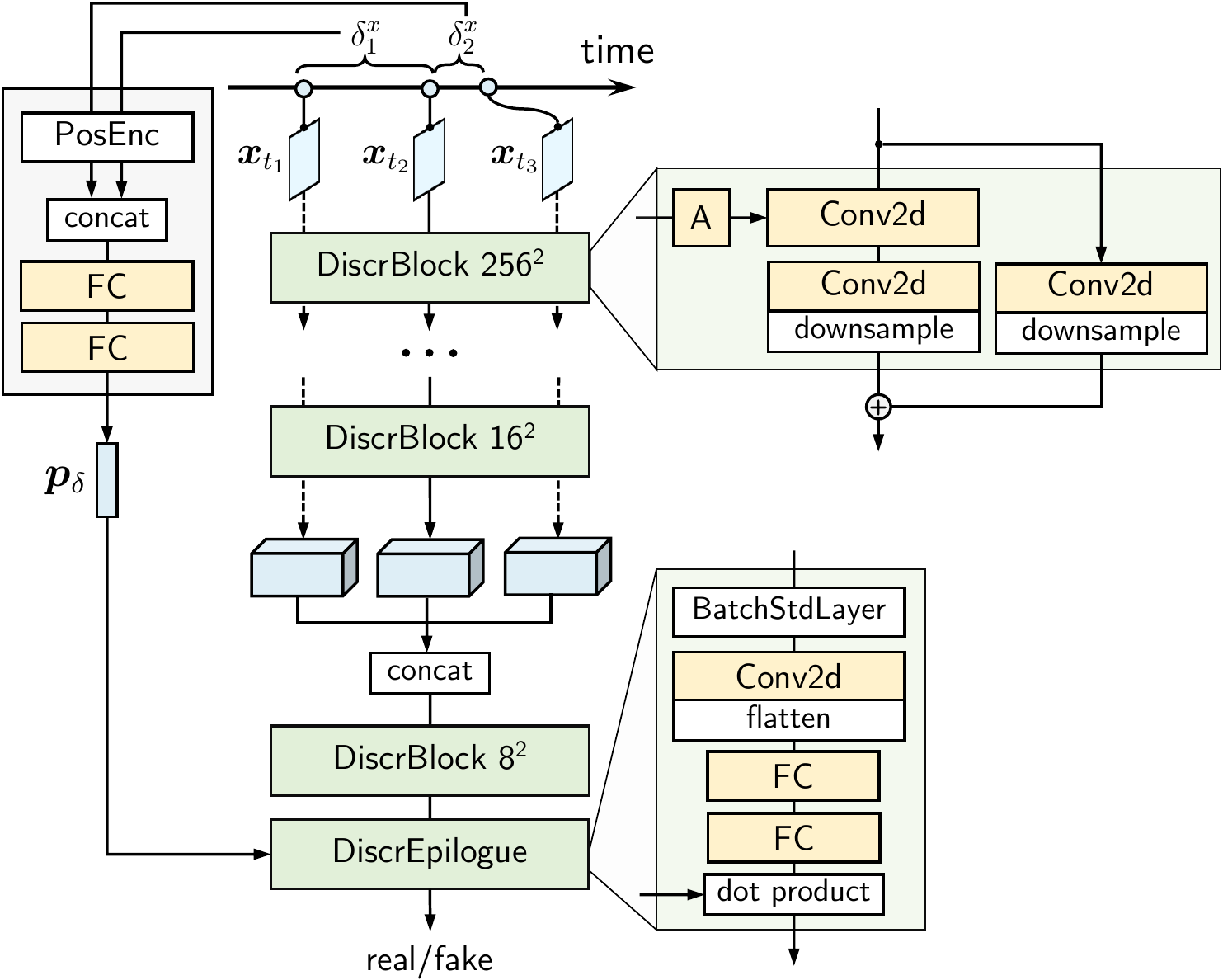}
    \caption{Discriminator architecture for $k=3$ frames per video. The only changes we do on top of the StyleGAN2~\cite{StyleGAN2} discriminator are concatenating activations channel-wise at the $16^2$ resolution and conditioning the model on the positional embeddings of the time distances between frames.}
    \label{fig:discriminator}
\end{figure}

%% file: figures/samples.tex
\begin{figure*}
    \centering
    \begin{subfigure}[b]{0.99\textwidth}
        \centering
        \includegraphics[width=\textwidth]{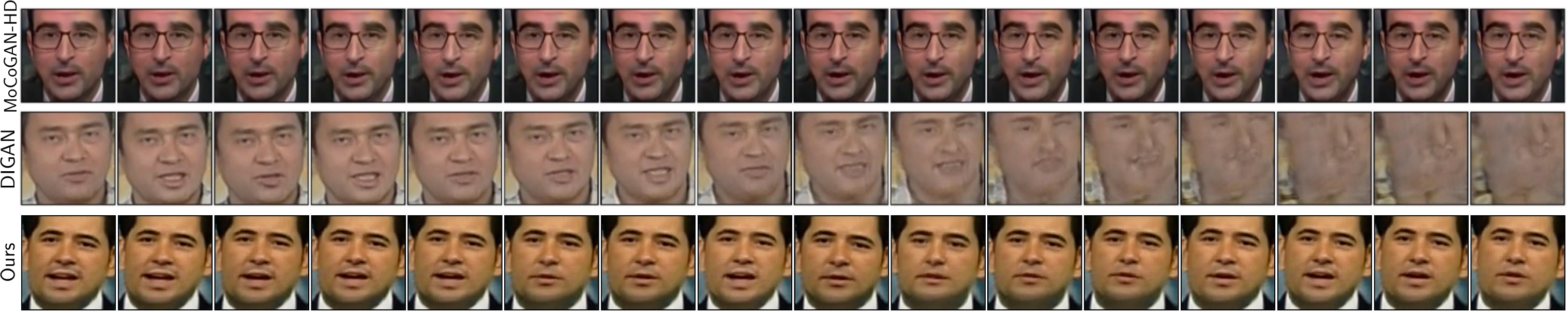}
    \end{subfigure}
    
    
    \begin{subfigure}[b]{0.99\textwidth}
        \centering
        \includegraphics[width=\textwidth]{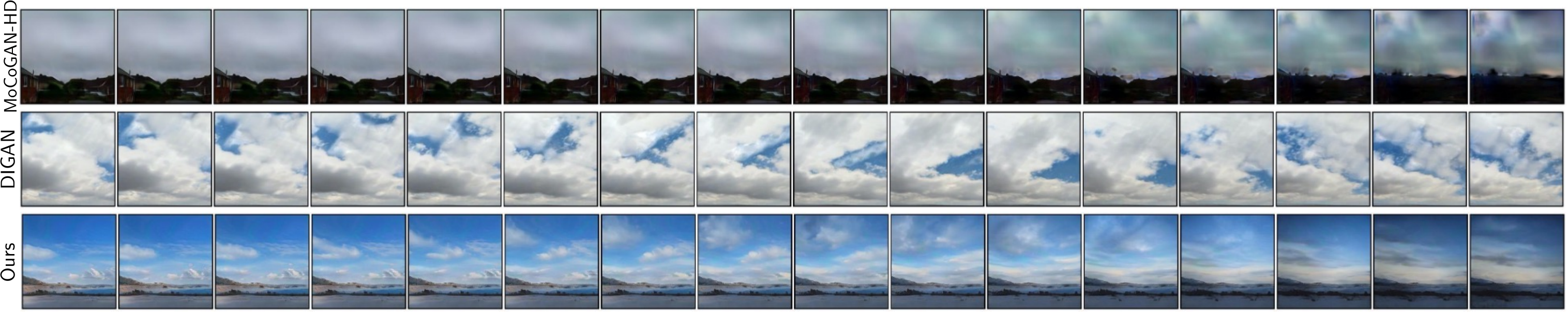}
    \end{subfigure}
    

    \begin{subfigure}[b]{0.99\textwidth}
        \centering
        \includegraphics[width=\textwidth]{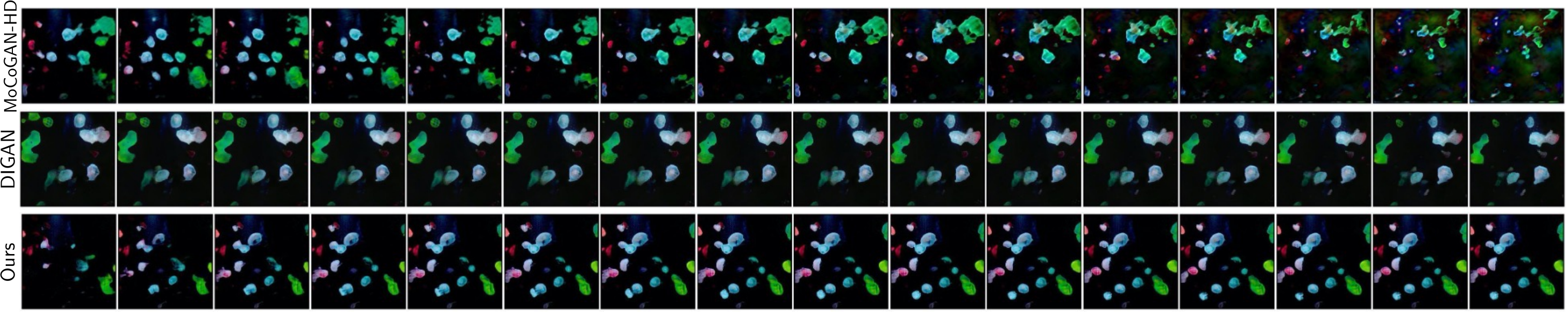}
    \end{subfigure}
    
    \caption{\textit{Random} samples from the existing methods on FaceForensics $256^2$, SkyTimelapse $256^2$ and RainbowJelly $256^2$, respectively. We sample a 64-frames video and display each 4-th frame, starting from $t = 0$.}
    \label{fig:samples}
\end{figure*}

%% file: sections/experiments.tex
\section{Experiments}\label{sec:experiments}

\textbf{Datasets}.
We test our model on 5 benchmarks: FaceForensics $256^2$~\cite{FaceForensics_dataset}, SkyTimelapse $256^2$~\cite{SkyTimelapse_dataset}, UCF101 $256^2$~\cite{UCF101_dataset}, RainbowJelly $256^2$ (introduced by us and described in Appx~\apref{ap:data}) and MEAD $1024^2$~\cite{MEAD_dataset}.
We used the \texttt{train} splits (when available) for all the datasets except for UCF101, where we used \texttt{train+test} splits.
We provide the datasets details in Appx~\apref{ap:data}.

\input{tables/main-results}

\textbf{Evaluation}.
Following prior work, we use Frechet Video Distance (FVD)~\cite{FVD} and Inception Score (IS) as our evaluation metrics with FVD being the main one since FID (its image-based counterpart) better aligns with human-perceived quality \cite{FID}.
We use two versions of FVD: FVD$_{16}$ and FVD$_{128}$, which use 16 and 128-frames-long videos to compute the statistics.
Inception Score is used only to evaluate the generation quality on UCF-101 since it uses a UCF-101-finetuned C3D model~\cite{TGAN}.

The official FVD implementation~\cite{FVD} does not provide a \textit{complete} evaluation pipeline, but rather an inference script for a single batch of videos, which are required to be \textit{already resized to $256^2$ and loaded into memory}.
This creates discrepancies in the evaluation protocols used by previous works since FVD (similar to FID~\cite{FID_evaluation}) is very sensitive to the subsampling and data processing procedures.
We implement, document (see Appx~\apref{ap:fvd}) and release a complete FVD evaluation protocol and use it to evaluate all the methods.

\textbf{Baselines}.
We use 5 baselines for comparison: MoCoGAN~\cite{MoCoGAN}, MoCoGAN~\cite{MoCoGAN} with the StyleGAN2~\cite{StyleGAN2} backbone, VideoGPT~\cite{VideoGPT}, MoCoGAN-HD~\cite{MoCoGAN-HD} and DIGAN~\cite{DIGAN}.
For MoCoGAN with the StyleGAN2 backbone (denoted as MoCoGAN-SG2), we replaced its generator and image-based discriminator with the corresponding StyleGAN2's components, leaving its video discriminator unchanged.
We also used the training scheme and regularizations from StyleGAN2.
MoCoGAN was trained for 5 days on a single GPU since its lightweight DC-GAN\cite{DC_GAN} backbone makes it fast to train, while MoCoGAN+SG2 was trained for 2 days on $\times 4$ GPUs to reach 25M real images seen by its image-based discriminator.
MoCoGAN-HD is trained for ${\approx}$4.5 days on $\times 4$ v100 GPUs, as specified in the original paper (Appx B of \cite{MoCoGAN-HD}).
We trained VideoGPT for the maximum affordable total time of 32 GPU-days in our resource constraints.
DIGAN~\cite{DIGAN} was trained for ${\approx}4$ days since after that its FVD score either did not change or exploded (on RainbowJelly).
We also replaced its weighted sampling strategy (selecting clips from longer videos with higher probabilities) with the uniform one, which is used by other methods~\cite{MoCoGAN, MoCoGAN-HD, VideoGPT}.
For each method, we used the checkpoint with the lowest FVD$_{16}$ value.

\subsection{Main experiments}\label{sec:experiments:main}
For the main evaluation, we train our method and all the baselines from scratch on the described $256^2$ datasets.
Each model is trained on $\times 4$ NVidia V100 32 GB GPUs, except for VideoGPT, which is very demanding in terms of GPU memory for $256^2$ resolution and we had to train it on $\times 4$ NVidia A6000 instead (with the overall batch size of \textit{4}).
For our method and MoCoGAN+SG2, we use exactly the same optimization scheme as StyleGAN2, including the loss function, Adam optimizer hyperparameters and R1 regularization~\cite{R1_reg}.
We reduce the learning rate by 10 for the $D_V$ module of MoCoGAN+SG2 since it does not have equalized learning rate~\cite{ProGAN}.
We use $\delta^z = 16$ for all the experiments except for SkyTimelapse, where we used $\delta^z = 256$.
See other training details in Appx~\apref{ap:training-details}.
We evaluate all the methods under the same evaluation protocol, described in Appx~\apref{ap:fvd} and report the results in Table~\ref{table:main-results}.

To measure the efficiency, we use the amount of GPU days required to train a method.
We build on top of the official StyleGAN2 implementation.\footnote{\href{https://github.com/NVlabs/stylegan2-ada-pytorch}{https://github.com/NVlabs/stylegan2-ada-pytorch}}
The training cost of the image-based StyleGAN2 to reach its specified 25M images is $7.72$ NVidia V100 GPU-days in our environment.
\modelname\ is trained for 2 days, which corresponds to ${\approx}$23M real frames seen by the discriminator.
MoCoGAN-HD is built on top of \texttt{stylegan2-pytorch}'s codebase\footnote{\href{https://github.com/rosinality/stylegan2-pytorch}{https://github.com/rosinality/stylegan2-pytorch}}, which is ${\approx}2$ times slower than the highly optimized NVidia's implementation.
That's why in Table~\ref{table:main-results} we report its training cost \textit{reduced} by a factor of 2 to account for this.


Our method significantly outperforms the existing ones on almost all the benchmarks in terms FVD$_{16}$ and FVD$_{128}$.
We visualize the samples in Fig~\ref{fig:long-videos} and Fig~\ref{fig:samples}.
Our method is able to generate hour-long plausibly looking videos, though the motion diversity and global motion coherence for them would be limited (see Appx~\apref{ap:limitations}).
MoCoGAN-HD suffers from the LSTM instability when unrolled to large lengths and does not produce diverse motions.
DIGAN produces high-quality videos on SkyTimelapse because its inductive bias of having joint spatio-temporal positional information is well suited for videos that have an entire scene moving.
But for FaceForensics, this leads to a ``head flying away'' effect (see Appx~\apref{ap:digan}).
To generate 1-hour long videos from MoCoGAN-HD, we unroll its LSTM model to the required depth (${\approx}90k$ steps) and synthesize frames only in the necessary time positions, while DIGAN, similar to our method, is able to generate frames non-autoregressively.

\subsection{Ablations}\label{sec:experiments:ablations}
To ablate the core components, we replaced $\G$ or $\D$ modules with their MoCoGAN+SG2 counterparts.
In the both cases, their removal leads to poor short-term and long-term video quality, as specified by the corresponding metrics in Table~\ref{table:ablations} and video samples in the supplementary.

Replacing continuous motion codes $\bm v(t)$ with $\bm u(t)$, produced by LSTM hurts the performance, especially when the distance $\delta^z$ between motion codes is small.
This happens due to unnaturally abrupt transitions between frames and we provide the corresponding samples in the supplementary.
The corresponding results are in Table~\ref{table:ablations}.

We also verify the importance of the conditioning in $\D$ and denote the experiment where it's disabled as ``w/o time conditioning'' in Table~\ref{table:ablations}.
Removing the time conditioning hurts the performance, because it constrains the ability of $\D$ to understand the temporal scale it is currently operating on.

An important design choice is how many samples per video one should use during training.
We try different values of $k$ for $k = 2, 3, 4, 8$ and $16$ and report the corresponding results in Table~\ref{table:sampling}.
As being discussed in \S\ref{sec:method:sampling}, for existing video generation benchmarks, it might be enough to sample only several frames per each video, and our experiments confirm this observation.
The performance is decreased for larger $k$, but this might be attributed to a weaker temporal aggregation procedure of $\D$, which simply concatenates features together.
It is surprising to see that modern datasets can be fit with as few as 2 samples per video.

\subsection{Properties}\label{sec:experiments:properties}

\input{figures/long-videos-mead}

\textit{Our generator is able to generate arbitrarily long videos}.
Our design of motion codes allows \modelname\ not to suffer from stability problems when unrolled to large (potentially infinite) video lengths.
This is verified by visualizing the video clips for the extremely large timesteps in Fig~\ref{fig:long-videos} and Fig~\ref{fig:long-videos-mead}.
We also demonstrate its ability to produce videos in arbitrarily high frame-rate in the supplementary.

\textit{Our model has the same latent space manipulation properties as StyleGAN2}.
To show this, we conduct two experiments: embedding, editing and animating an off-the-shelf image and editing and animating the first frame of a generated video.
To embed an image, we used the optimization procedure similar to \cite{Image2StyleGAN}, but considering it to be positioned at $t = 0$.
To edit an image with CLIP, we used the procedure of \cite{StyleCLIP}.
The results of these experiments are visualized in Fig~\ref{fig:latent-manipulation} and we provide the details in Appx~\apref{ap:training-details} and more examples in the supplementary.
Apart from showing the good properties of its latent space, these experiments demonstrate the extrapolation potential of our generator.

\textit{\modelname\ has almost the same training efficiency and image quality as StyleGAN2}.
In Fig~\ref{fig:compare-to-sg2}, we plot the FID scores (computed from 16-frames videos) and training costs of modern video generators on FaceForensics $256^2$ by their corresponding FVD$_{16}$ scores.
Our method comes very close to StyleGAN2: it converges to FID of 9.44 in 8 GPU-days compared to FID of 8.42 in 7.72 GPU-days for StyleGAN2, which is only ${\approx}$10\% worse.
This raises the question whether video generators can be as computationally efficient and good in terms of image quality as image ones.

\textit{Our model is the first one which is directly trainable on $1024^2$ resolution}.
We provide the generations on MEAD $1024^2$ for our method and for MoCoGAN-HD.
MoCoGAN-HD cannot preserve the identity of a speaker and diverges for large video lengths, while our method achieves comparable image quality and coherent motions.
For this dataset, our model was trained for 7 days on $\times 4$ NVidia v100 GPUs and obtained FID of 24.12 and FVD$_{16}$ of 156.1.
Image generator for MoCoGAN-HD was trained for $14$ days on $\times 4$ A6000 GPUs, while its video generator was trained for only $5$ days since it didn't require high-resolution training.

\textit{Our discriminator provides more informative learning signal to $\G$}.
Fig~\ref{fig:grads-vis} visualizes the gradient signal to the generator from our discriminator and the \texttt{conv3d}-based video discriminator of MoCoGAN-HD, measured at ${\approx}50\%$ of training for our method (at 10M images seen by $\D$) and MoCoGAN-HD (at the 300-th epoch).
In our case, one can easily see fine-grained details of the face structure, perceived by $\D$, while in case of MoCoGAN-HD, most of the gradient is redundant and lack any structural information.

\textit{Content and motion decomposition.}
Similar to MoCoGAN~\cite{MoCoGAN}, our generator captures content and motion variations in a disentangled manner: altering motion codes $\zm_{t_0}, ..., \zm_{t_n}$ while fixing $\zc$ does not change the appearance variations (like, a speaker's identity).
Similarly, re-sampling $\zc$ does not influence motion patterns on a video, but only its content.
We provide the corresponding visualizations on the project website.

\input{tables/ablations}
\input{tables/sampling-ablations}

%% file: tables/main-results.tex
\begin{table*}[]
\caption{Quantitative performance and training cost of different methods.
We trained \textit{all} the methods from scratch on $256^2$ resolution datasets using the official codebases and evaluated them under the unified evaluation protocol (see \S\ref{sec:experiments}). Training was done on $\times 4$ 32 GB NVidia V100 GPUs for all the methods except VideoGPT, which was trained on $\times 4$ NVidia A6000 GPUs (with 48.5 GB of memory each) due to its high memory consumption. For 2-stage methods, we report their training cost in the ``$X+Y$'' format. $^\dagger$VideoGPT was trained for our maximum resource constraint of 32 GPU-days which was detrimental to its performance on $256^2$ resolution. Vanilla StyleGAN2 training time on $256^2$ resolution (with mixed precision and optimizations~\cite{StyleGAN2-ADA}) is 7.72 GPU-days in our environment.}
\label{table:main-results}
\centering
\resizebox{1.0\linewidth}{!}{
\begin{tabular}{lccccccccc}
\toprule
\multirow{2}{*}{Method} & \multicolumn{2}{c}{FaceForensics} & \multicolumn{2}{c}{SkyTimelapse} & \multicolumn{2}{c}{UCF101} & \multicolumn{2}{c}{RainbowJelly} & \multirowcell{2}{Training cost \\ (GPU-days)} \\
& FVD$_{16}$ & FVD$_{128}$ & FVD$_{16}$ & FVD$_{128}$ & FVD$_{16}$ & FVD$_{128}$ & FVD$_{16}$ & FVD$_{128}$ \\
\midrule
MoCoGAN \cite{MoCoGAN} & 124.7 & 257.3 & 206.6 & 575.9 & 2886.9 & 3679.0 & 1572.9 & 549.7 & \textbf{5} \\
~+ StyleGAN2 backbone & 55.62 & 309.3 & 85.88 & 272.8 & 1821.4 & 2311.3 & 638.5 & 463.0 & 8 \\
MoCoGAN-HD \cite{MoCoGAN-HD} & 111.8 & 653.0 & 164.1 & 878.1 & 1729.6 & 2606.5 & 579.1 & 628.2 & 7.5 + 9 \\
VideoGPT \cite{VideoGPT}$^\dagger$ & 185.9 & N/A & 222.7 & N/A & 2880.6 & N/A & \textbf{136.0} & N/A & 16 + 16 \\
DIGAN \cite{DIGAN} & 62.5 & 1824.7 & 83.11 & \textbf{196.7} & 1630.2 & 2293.7 & 436.6 & 369.0 & 16 \\
\modelname\ \ours & \textbf{47.41} & \textbf{89.34} & \textbf{79.52} & 197.0 & \textbf{1431.0} & \textbf{1773.4} & 195.4 & \textbf{262.5} & 8 \\
\bottomrule
\end{tabular}
}
\end{table*}

%% file: figures/long-videos-mead.tex
\begin{figure}
    \centering
    \includegraphics[width=\linewidth]{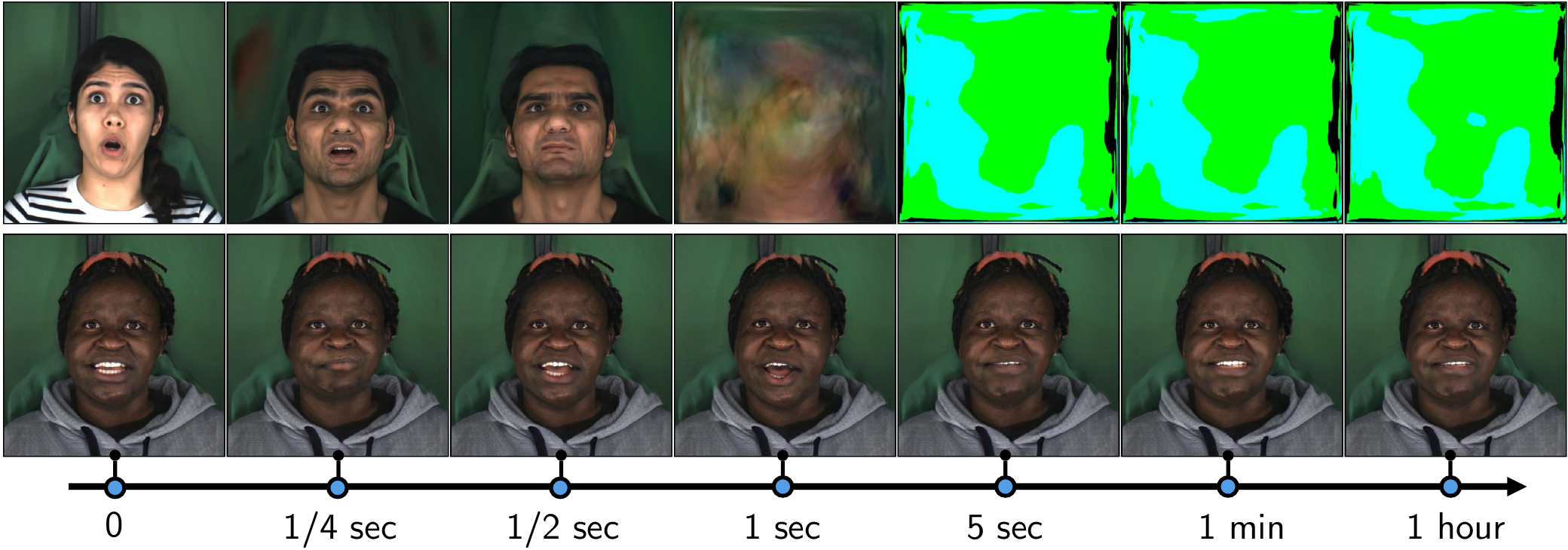}
    \caption{Generations on MEAD $1024^2$ \cite{MEAD_dataset} for MoCoGAN-HD \cite{MoCoGAN-HD} and our method. MoCoGAN-HD cannot preserve the identity and diverges for long LSTM unrolling. (Note that all videos in the dataset have static head positions --- see Appx~\apref{ap:data}).}
    \label{fig:long-videos-mead}
    \vspace{-0.5cm}
\end{figure}

%% file: tables/ablations.tex
\begin{table}[]
\caption{Ablating architectural components of our model.}
\label{table:ablations}
\centering
\resizebox{1.0\linewidth}{!}{
\begin{tabular}{lcccc}
\toprule
\multirow{2}{*}{Method} & \multicolumn{2}{c}{FaceForensics $256^2$} & \multicolumn{2}{c}{SkyTimelapse $256^2$} \\
& FVD$_{16}$ & FVD$_{128}$ & FVD$_{16}$ & FVD$_{128}$ \\
\midrule
Default \modelname\ & 47.41 & 89.34 & 79.52 & 197.0 \\
\midrule
w/o our $\G$ & 65.88 & 41.77 & 109.1 & 240.2 \\
w/o our $\D$ & 154.0 & 139.1 & 236.9 & 258.0 \\
\midrule
w/o time conditioning in $\D$ & 95.4 & 236.0 & 102.1 & 210.3 \\
\midrule
w LSTM codes, $\delta^z = 1$ & 131.9 & 159.1 & 135.7 & 196.1 \\
w LSTM codes, $\delta^z = 16$ & 180.3 & 94.55 & 95.71 & 165.8 \\
\bottomrule
\end{tabular}
}
\end{table}

%% file: tables/sampling-ablations.tex
\begin{table}[]
\caption{Ablating the amount of frames $k$ per clip used during training. Sparse training provides better results for our method.}
\label{table:sampling}
\centering
\resizebox{1.0\linewidth}{!}{
\begin{tabular}{lcccc}
\toprule
\multirow{2}{*}{Number of frames} & \multicolumn{2}{c}{FaceForensics $256^2$} & \multicolumn{2}{c}{SkyTimelapse $256^2$} \\
& FVD$_{16}$ & FVD$_{128}$ & FVD$_{16}$ & FVD$_{128}$ \\
\midrule
$k=2$ & 60.41 & 93.5 & 50.5 & 209.9 \\
$k=3$ (default) & 47.41 & 89.34 & 79.52 & 197.0 \\
$k=4$ & 51.84 & 114.9 & 65.7 & 194.5 \\
$k=8$ & 101.9 & 211.4 & 73.12 & 215.9 \\
$k=16$ & 92.52 & 192.8 & 107.6 & 254.3 \\
\bottomrule
\end{tabular}
}
\end{table}

%% file: sections/conclusion.tex
\section{Conclusion}

In this work, we provided a different perspective on time for video synthesis and built a continuous video generator using the paradigm of neural representations.
For this, we developed motion representations through the lens of positional embeddings, explored sparse training of video generators and redesigned a typical dual structure of a video discriminator.
Our model is built on top of StyleGAN2 and features a lot of its perks, like efficient training, good image quality and editable latent space.
We hope that our work would serve as a solid basis for building more powerful video generators in the future.
The limitations and potential negative impact are discussed in Appx~\apref{ap:limitations}.

%% file: appendix/limitations.tex
\section{Limitations and potential negative impact}\label{ap:limitations}

\subsection{Limitations}
Our model has the following limitations:
\begin{itemize}
    \item \emph{Limitations of sparse training}. In general, sparse training makes it impossible for $\D$ to capture complex dependencies between frames. But surprisingly, it provides state-of-the-art results on modern datasets, which (using the statement from \S\ref{sec:method:sampling}) implies that they are not that sophisticated in terms of motion.
    \item \emph{Dataset-induced limitations}. Similar to other machine learning models, our method is bound by the dataset quality it is trained on. For example, for FaceForensics $256^2$ dataset~\cite{FaceForensics_dataset}, our embedding and manipulations results are inferior to StyleGAN2 ones~\cite{Image2StyleGAN}. This is due to the limited number of identities (just ~700) in FaceForensics and their larger diversity in terms of quality compared to FFHQ~\cite{StyleGAN}, which StyleGAN2 was trained on.
    \item \emph{Periodicity artifacts}. $\G$ still produces periodic motions sometimes, despite of our acyclic positional embeddings. Future investigation on this phenomena is needed.
    \item \emph{Poor handling of new content appearing}. We noticed that our generator tries to reuse the content information encoded in the global latent code as much as possible. It is noticeable on datasets where new content appears during a video, like Sky Timelapse or Rainbow Jelly. We believe it can be resolved using ideas similar to ALIS~\cite{ALIS}.
    \item \emph{Sensitivity to hyperparameters}. We found our generator to be sensitive to the minimal initial period length $\max_i \sigma_i$ (See Appx~\apref{ap:training-details}). We increased it for SkyTimelapse~\cite{SkyTimelapse_dataset} from 16 to 256: otherwise it contained unnatural sharp transitions.
\end{itemize}
We plan to address those limitations in our future works.

\subsection{Potential negative impact}
The potential negative impact of our method is similar to those of traditional image-based GANs: creating ``deepfakes'' and using them for malicious purposes.\footnote{\href{https://en.wikipedia.org/wiki/Deepfake}{https://en.wikipedia.org/wiki/Deepfake}}.
Our model made it much easier to train a model which produces much more realistic video samples with a small amount of computational resources.
But since the availability of high-quality datasets is very low for video synthesis, the resulted model will fall short compared to its image-based counterpart, which could use rich, extremely qualitative image datasets for training, like FFHQ~\cite{StyleGAN}.

%% file: appendix/training-details.tex
\section{Implementation and training details}\label{ap:training-details}

Note, that all the details can be found in the source code: \href{\codeurl}{\codeurl}.

\subsection{Optimization details and hyperparameters}

Our model is built on top of the official StyleGAN2-ADA~\cite{StyleGAN2-ADA} repository\footnote{\href{https://github.com/nvlabs/stylegan2-ada}{https://github.com/nvlabs/stylegan2-ada}}.
In this work, we build a model to generate continuous videos and a reasonable question to ask was why not use INR-GAN~\cite{INR-GAN} instead (like DIGAN~\cite{DIGAN}) to have fully continuous signals?
The reason why we chose StyleGAN2 instead of INR-GAN is that StyleGAN2 is amenable to the mixed-precision training, which makes it train ${\approx}2$ times faster.
For INR-GAN, enabling mixed precision severely decreases the quality and we hypothesize the reason if it is that each pixel in INR-GAN's activations tensor carries more information (due to the spatial independence) since the model cannot spatially distribute information anymore.
And explicitly restricting the range of possible values adds a strict upper bound on the amount of information one each pixel is able to carry.
We also found that adding coordinates information does not improve video quality for our generator neither qualitatively, nor in terms of scores.

Similar to StyleGAN2, we utilize non-saturating loss and $R_1$ regularization with the loss coefficient of 0.2 in all the experiments, which is inherited from the original repo and we didn't try any hyperparameter search for it.
We also use the \texttt{fmaps} parameter of 0.5 (the original StyleGAN2 used fmaps parameter of 1.0), which controls the channel dimensionalities in $\G$ and $\D$, since it is the default setting for StyleGAN2-ADA for $256^2$ resolution.
This allowed us to further speedup training.

The dimensionalities of $\bm w, \bm z, \bm u_t, \bm v_t$ are all set to 512.

As being stated in the main text, we use a padding-less \texttt{conv1d}-based motion mapping network $\Fm$ with a large kernel size to generate raw motion codes $\bm u_t$.
In all the experiments, we use the kernel size of $11$ and stride of $1$.
We do not use any dilation in it despite the fact that they could increase the temporal receptive field: we found that varying the kernel size didn't produce much benefit in terms of video quality.
Using padding-less convolutions allows the model to be stable when unrolled at large depths.
We use 2 layers of such convolutions with a hidden size of 512.
Another benefit of using \texttt{conv1d}-based blocks is that in contrast to LSTM/GRU cells one can practically incorporate equalized learning rate~\cite{ProGAN} scheme into it.

Using \texttt{conv1d}-based motion mapping network without paddings forces us to use ``previous'' motion noise codes $\zm_t$.
That's why instead of sampling a sequence $\zm_{t_0}, ..., \zm_{t_n}$, we sample a slightly larger one to adjust for the reduced sequence size.
For the same-padding strategy, for sampling a frame at position $t \in [t_{n-1}, t_n)$, we would need to produce $n$ motion noise codes $\zm$.
But with our kernel size of 11, with 2 layers of convolutions and without padding, the resulted sequence size is $n + 20$.

The training performance of VideoGPT on UCF101 is surprisingly low despite the fact that it was developed for such kind of datasets~\cite{VideoGPT}. 
We hypothesize that this happens due to UCF101 being a very difficult dataset and VideoGPT being trained with the batch size of 4 (higher batch size didn't fit our 200 GB GPU memory setup), which damaged its ability to learn the distribution.

To train our model, we also utilized adaptive differentiable augmentations of StyleGAN2-ADA~\cite{StyleGAN2-ADA}, but we found it important to make them \emph{video-consistent}, i.e. applying the same augmentation for each frame of a video.
Otherwise, the discriminator starts to underperform, and the overall quality decreases.
We use the default \texttt{bgc} augmentations pipe from StyleGAN2-ADA, which includes horizontal flips, 90 degrees rotations, scaling, horizontal/vertical translations, hue/saturation/brightness/contrast changes and luma axis flipping.

While training the model, for real videos we first select a video index and then we select random clip (i.e., a clip with a random offset).
This differs from the traditional DIGAN or VideoGPT training scheme, that's why we needed to change the data loaders to make them learn the same statistics and not get biased by very long videos.

To develop this project, ${\approx}7.5$ NVidia v100 32GB GPU-years + ${\approx}0.3$ NVidia A6000 GPU-years were spent.

\subsection{Projection and editing procedures}
In this subsection, we describe the embedding and editing procedures, which were used to obtain results in Fig~\ref{fig:latent-manipulation}.

\textbf{Projection}.
To project an existing photogrpah into the latent space of $\G$, we used a procedure from StyleGAN2~\cite{StyleGAN2}, but projecting into $\mathcal{W}+$ space~\cite{Image2StyleGAN} instead of $\mathcal{W}$, since it produces better reconstruction results and does not spoil editing properties.
We set the initial learning rate to $0.1$ and optimized a $\bm w$ code for LPIPS reconstruction loss~\cite{LPIPS} for 1000 steps using Adam.
For motion codes, we initializated a static sequence and kept it fixed during the optimization process.
We noticed that when it is also being optimized, the reconstruction becomes almost perfect, but it breaks when another sequence of motion codes is plugged in.

\textbf{Editing}.
Our CLIP editing procedure is very similar to the one in StyleCLIP~\cite{StyleCLIP}, with the exception that we embed an image assuming that it is a video frame in location $t = 0$.
On each iteration, we resample motion codes since all our edits are semantic and do not refer to motion.
We leave the motion editing with CLIP for future exploration.
For the sky editing video presented in Fig~\ref{fig:latent-manipulation}, we additionally utilize masking: we initialize a mask to cover the trees and try not to change them during the optimization using LPIPS loss.
For all the videos, presented in the supplementary website, \emph{no masking is used}.




The details can be found in the provided source code.

\subsection{Additional details on positional embeddings}

\textbf{Mitigating high-frequency artifacts}.
We noticed that if our periods $\bm\omega_t$ are left unbounded, they might grow to very large values (up to magnitude of ${\approx}20.0$), which corresponds to extra high frequencies (the period length becomes less than 4 frames) and leads to temporal aliasing.
That's why we process them via the $\text{tanh}(\bm\omega_t) + 1$ transform: this bounds them into $(0, 2)$ range with the mean of 1.0, i.e. using the at-initialization frequency scaling, which we discuss next.

\textbf{Linearly spaced periods}.
An important design decision is the scaling of periods since at initialization it should cover both high-frequency and low-frequency details.
Existing works use either exponential scaling $\bm\sigma = (2\pi/2^{d}, 2\pi/2^{d-1}, ...)$ (e.g., \cite{NeRF, Nerfies, ALIS, SAPE}) or random scaling $\bm\sigma \sim \mathcal{N}(0, \xi \bm{I})$ (e.g., \cite{SIREN, FourierFeatures, INR-GAN, CIPS}).
In practice, we scale the $i$-th column of the amplitudes weight matrix with the value:
\begin{equation}\label{eq:freq-scaling}
\sigma_i = \frac{2\pi}{\omega_\text{min} + (i/N) \cdot (\omega_\text{max} - \omega_\text{min})},
\end{equation}
where we use $\omega_\text{max} = 2^{10}$ frames and $\omega_\text{min} = 2^{3}$ frames in all the experiments, except for SkyTimelapse, for each we use $\omega_\text{min} = 2^8$.
We call this scheme \emph{linear scaling} and use it as an additional tool to alleviate periodicity since it greatly increases the overall cycle of a positional embedding (see Fig~\ref{fig:time-embs}).
See also the accompanying source code for details.


\input{figures/time-embs}

Another benefit of using our positional embeddings over LSTM is that they are ``always stable'', i.e. they are always in a suitable range.

%% file: figures/time-embs.tex
\newcommand{\timeEmbsSubfigureScale}{1.0}
\newcommand{\timeEmbsSubfigureVspace}{-0.3cm}

\begin{figure}
    \centering
    \begin{subfigure}[b]{\timeEmbsSubfigureScale\linewidth}
        \centering
        \includegraphics[width=\textwidth]{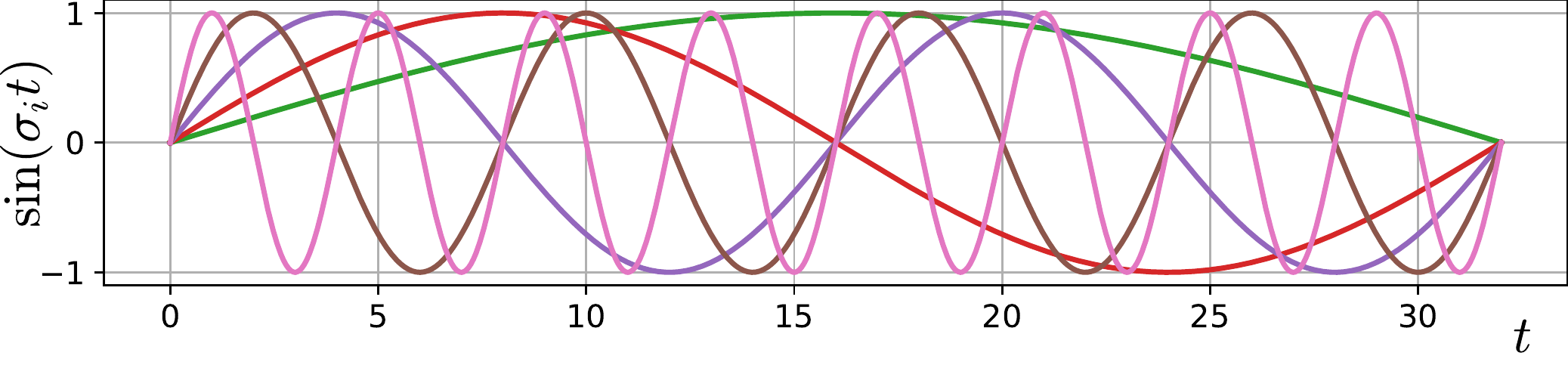}
        \caption{Exponentially spaced periods~\cite{NeRF}: $d=5$, cycle length is 64.}
        \label{fig:time-embs:exp-spaced}
    \end{subfigure}
    
    \vspace{\timeEmbsSubfigureVspace}
    \hfill
    
    \centering
    \begin{subfigure}[b]{\timeEmbsSubfigureScale\linewidth}
        \centering
        \includegraphics[width=\textwidth]{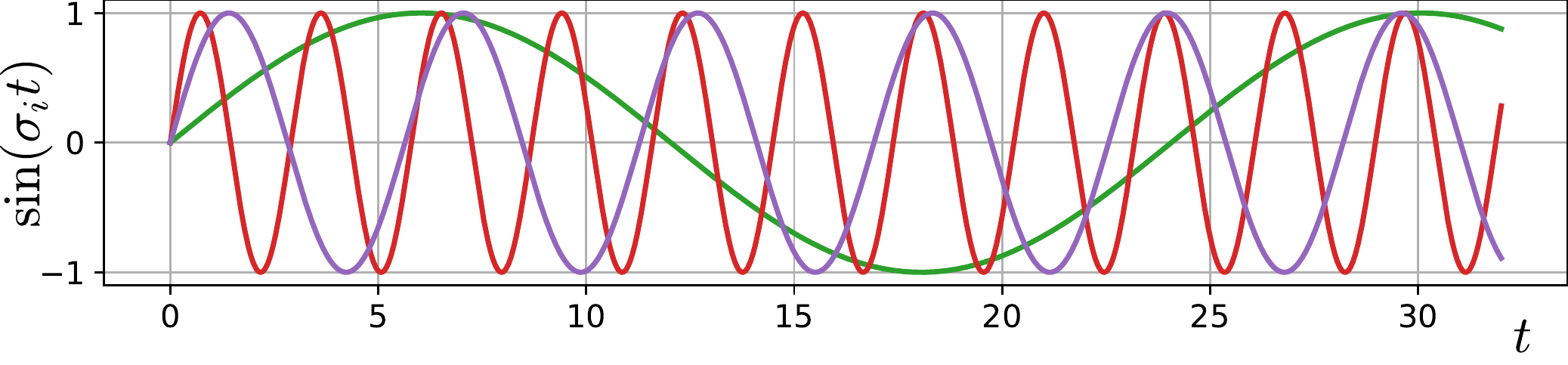}
        \caption{Random periods~\cite{SIREN, FourierFeatures}: $d=3$, cycle length is 120 (for the depicted $\bm \sigma \sim \Nstd$).}
        \label{fig:time-embs:random}
    \end{subfigure}
    
    \vspace{\timeEmbsSubfigureVspace}
    \hfill
    
    \centering
    \begin{subfigure}[b]{\timeEmbsSubfigureScale\linewidth}
        \centering
        \includegraphics[width=\textwidth]{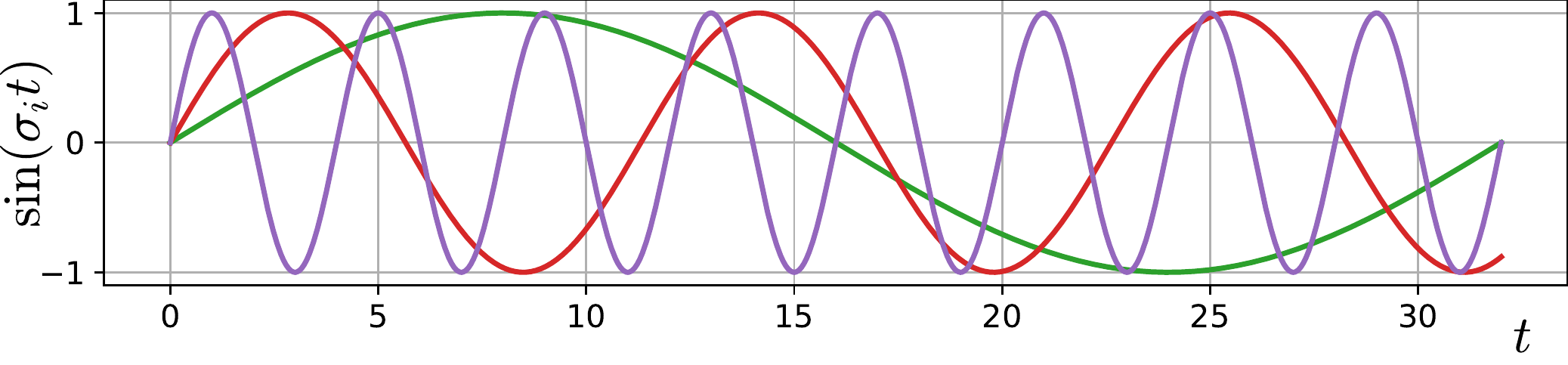}
        \caption{Linearly spaced periods \ours: $d=3$, cycle length is 352.}
        \label{fig:time-embs:linearly-spaced}
    \end{subfigure}
    
    \vspace{\timeEmbsSubfigureVspace}
    \hfill
    
    \centering
    \begin{subfigure}[b]{\timeEmbsSubfigureScale\linewidth}
        \centering
        \includegraphics[width=\textwidth]{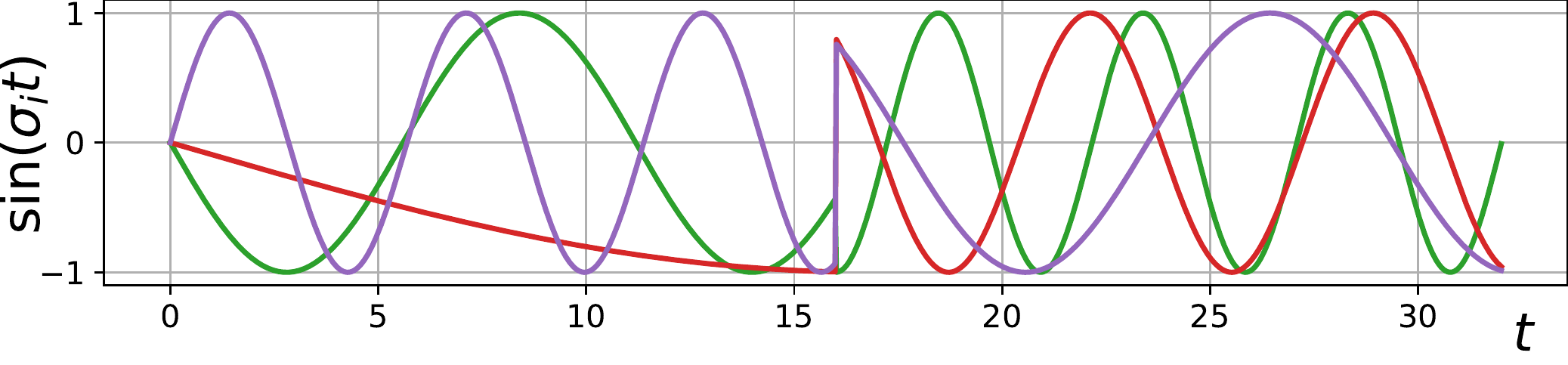}
        \caption{Raw acyclic positional embeddings $\tilde{\bm v_t}$: $d = 3$, no cyclicity. While such embeddings are acyclic, they have discontinuities at stitching points.}
        \label{fig:time-embs:ours-raw}
    \end{subfigure}
    
    \vspace{\timeEmbsSubfigureVspace}
    \hfill
    
    \centering
    \begin{subfigure}[b]{\timeEmbsSubfigureScale\linewidth}
        \centering
        \includegraphics[width=\textwidth]{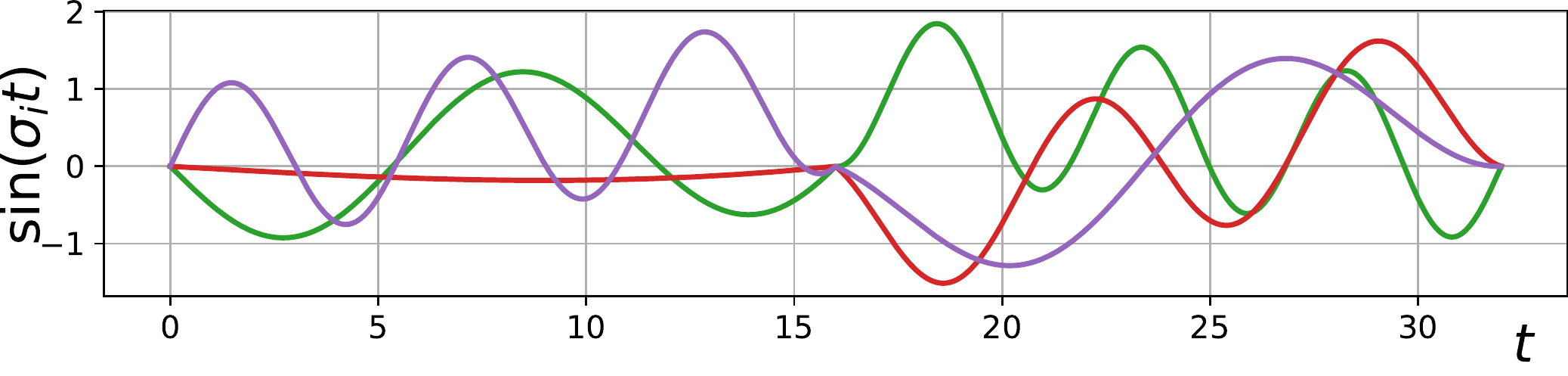}
        \caption{Stitched raw acyclic positional embeddings without alignment vectors: $d=3$, no cyclicity. Stitching raw positional embeddings without using ``aligners'' $\bm a = W_a \bm u$ removes discontinuities, but reduces the expressive power of positional embeddings since they have zero values at time locations $\{t_0, t_1, ..., t_n, ...\}$.}
        \label{fig:time-embs:ours-no-aligners}
    \end{subfigure}
    
    \vspace{\timeEmbsSubfigureVspace}
    \hfill
    
    \centering
    \begin{subfigure}[b]{\timeEmbsSubfigureScale\linewidth}
        \centering
        \includegraphics[width=\textwidth]{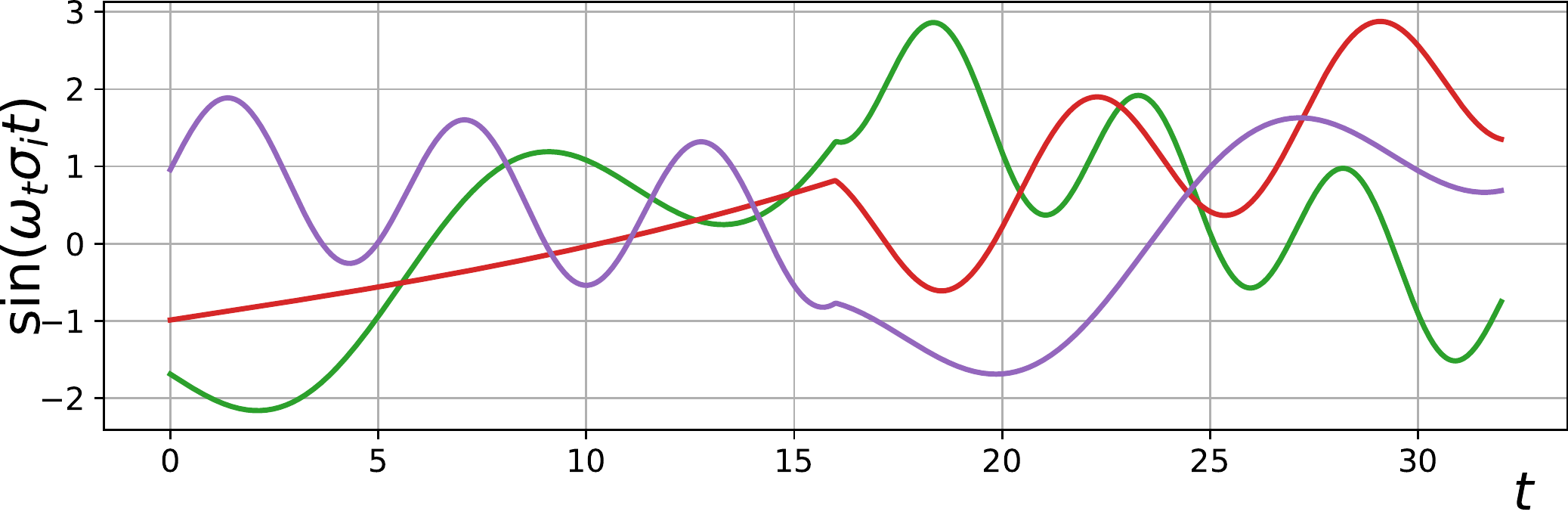}
        \caption{Acyclic periods with linearly-spaced scaling \ours: $d=3$, no cyclicity. Notice that the frequencies and phases are controlled by the motion mapping network $\Fm$: for example, it has the possibility to accelerate some motion (like the one represented by the red curve) by increasing its frequency.}
        \label{fig:time-embs:ours}
    \end{subfigure}
    \caption{Visualizing positional embeddings $\sin(\bm\sigma t)$ for different initialization strategies of periods scales $\bm\sigma$. The cycle length is the minimum value of $t$ for which the positional embedding vector starts repeating itself (it is computed as a least common multiple of all the individual periods lengths). Existing works use either exponentially spaced or random scaling, but in our case we use the linearly spaced one since it has a very large global cycle (in contrast to exponential scaling) and is guaranteed to include high-frequency, medium-frequency and high-freqency waves (in contrast to random scaling).}
    \label{fig:time-embs}
\end{figure}

%% file: appendix/evaluation.tex
\section{Evaluation details}\label{ap:fvd}
For the practical implementation, see the provided source code: \href{\codeurl}{\codeurl}.

In this section, we describe the difficulties of a fair comparison of the FVD score.
There are discrepancies between papers in computing even FID \cite{FID_evaluation}.
So, it is less surprising that computing FVD for videos diverge even more and has even more implications for methods evaluation.

First, we note that I3D model \cite{i3d} has different weights on tf.hub \href{https://tfhub.dev/deepmind/i3d-kinetics-400/1}{https://tfhub.dev/deepmind/i3d-kinetics-400/1} --- the model which is used in the official FVD repo.\footnote{\href{https://github.com/google-research/google-research/blob/master/frechet\_video\_distance}{https://github.com/google-research/google-research/blob/master/frechet\_video\_distance}} --- compared to its official release in the official github repo implementation\footnote{\href{https://github.com/deepmind/kinetics-i3d}{https://github.com/deepmind/kinetics-i3d}}
That's why we manually exported the weights from tf.hub and used this github repo\footnote{\href{https://github.com/hassony2/kinetics_i3d_pytorch}{https://github.com/hassony2/kinetics\_i3d\_pytorch}} to obtain an exact implementation in Pytorch.

There are several issues with FVD metric on its own.
First, it does not capture motion collapse, which can be observed by comparing FVD$_{16}$ and FVD$_{128}$ scores between \modelname\ and \modelname with LSTM motion codes instead of our ones: the latter one has a severe motion collapse issue (see the samples on our website) and has similar or lower FVD$_{128}$ scores compared to our model: 196.1 or 165.8 (depending on the distance between anchors) vs 197.0 for our model.
Another issue with FVD calculation is that it is biased towards image quality.
If one trains a good image generator, i.e. a model which is not able to generate any videos at all, then FVD will still be good for it even despite the fact that it would have degenerate motion.

We also want to make a note on how we compute FID for vidoe generators.
For this, we generate 2048 videos of 16 frames each (starting with $t=0$) and use all those frames in the FID computation.
In this way, it gives ${\approx}$33k images to construct the dataset, but those images will have lower diversity compared to a typically utilized 50k-sized set of images from a traditional image generator~\cite{StyleGAN}.
The reason of it is that 16 images in a single clip likely share a lot of content.
A better strategy would be to generate 50k videos and pick a random frame from each video, but this is too heavy computationally for models which produce frames autoregressively.
And using just the first frame in FID computation will unfairly favour MoCoGAN-HD, which generates the very first frame of each video with a freezed StyleGAN2 model.

FVD is greatly influenced by 1) how many clips per video are selected; 2) with which offsets; and 3) at which frame-rate.
For example, SkyTimelapse contains several \textit{extremely} long videos: if we select as many clips as possible from each real video, that it will severely bias the statistics of FVD.
For FaceForensics, videos often contain intro frames during their first ${\approx}$0.5-1.0 seconds, which will affect FVD when a constant offset of $0$ is chosen to extract a single clip per video.

That's why we use the following protocol to compute FVD$_n$.

\textbf{Computing real statistics}.
To compute real statistics, we select a \textit{single} clip per video, chosen at a random offset.
We use the actual frame-rate of the dataset, which the model is being trained on, without skipping any frames.
The problem of such an approach is that for datasets with small number of long videos (like, FaceForensics, see Table~\ref{table:datasets-info}) might have noisy estimates.
But our results showed that the standard deviations are always $<3.0$ even for FaceForensics $256^2$.
The largest standard deviation we obserbed was when computing FVD$_{16}$ on RainbowJelly: on this dataset it was $26.15$ for VideoGPT, but it is $<1\%$ of its overall magnitude.

\textbf{Computing fake statistics}.
To compute fake statistics, we generate 2048 videos and save them as frames in JPEG format via the Pillow library.
We use the quality parameter $q=95$ for doing this, since it was shown to have very close quality to PNG, but without introducing artifacts that would lead to discrepancies~\cite{FID_evaluation}.
Ideally, one would like to store frames in the PNG format, but in this case it would be too expensive to represent video datasets: for example, MEAD $1024^2$ would occupy ${\approx}0.5$ terabytes of space in this case.

\input{tables/fvd}

We illustrate the subtleties of FVD computation in Table~\ref{table:fvd}.
For this, we compute real/fake statistics for our model in several different ways:
\begin{itemize}
    \item \textit{Resized to $128^2$}. Both fake and real statistics images are resized into $128^2$ resolution via the pytorch bilinear interpolation (without corners alignment) before computing FVD.
    \item \textit{JPG/PNG discrepancy}. Instead of saving fake frames in JPG with $q=95$, we use $q=75$ parameter in the PIL library. This creates more JPEG-like artifacts, which, for example, FID is very sensitive to.
    \item \textit{Using all clips per video}. We use all available $n$-frames-long clips in each video without overlaps. Note, that our model was trained 
    \item \textit{Using only first frames}. In each real video, instead of using random offsets to select clips, we use the first $n$ frames.
    \item \textit{Using $s=8$ subsampling}. When sampling frames for computing real/fake statistics, we select each $8$-th frame. This is the strategy which was employed for some of the experiments in the original paper~\cite{FVD} --- but in their case, authors trained the model on videos with this subsampling.
\end{itemize}

For completeness, we also provide the Inception Score~\cite{TGAN} on UCF-101 $256^2$ dataset in Table~\ref{table:inception-score}.
Note that is computed by resizing all videos to $112 \times 112$ spatial resolution (due to the internal structure of the C3D~\cite{c3d} model), which makes it impossible for it to capture high-resolution details of the generated videos, which is the focus of the current work.

\input{tables/inception-score}

In Tab~\ref{table:fid-scores}, we provide the numbers, used in Fig~\ref{fig:compare-to-sg2}.
Note that StyleGAN2 training in our case is slightly slower than the officially specified one (7.3 vs 7.7 GPU days)\footnote{\href{https://github.com/NVlabs/stylegan2-ada-pytorch}{https://github.com/NVlabs/stylegan2-ada-pytorch}}, which we attribute to a slightly slower file system on our computational cluster.

\input{tables/fid-scores}

%% file: tables/fvd.tex
\begin{table}[]
\caption{Subtleties of FVD calculation. We report different ways of calculating FVD$_{16}$ on FaceForensics $256^2$ (FF) and SktTimelapse $256^2$ (ST) for one of our checkpoints. We show how the scores of \modelname\ are influenced a lot when different strategies of FVD$_{16}$ calculation are employed. See the text for the description of each row.}
\label{table:fvd}
\centering
\resizebox{1.0\linewidth}{!}{
\begin{tabular}{lcc}
\toprule
Method & FF & ST \\
\midrule
Proper computation & 76.82 $\pm 1.57$ & 61.95 $\pm 0.92$ \\
When resized to $128^2$  & 38.92 & 59.86 \\
With jpg/png discrepancy & 80.17 & 71.40  \\
When using all clips per video & 84.59 & 72.03 \\
When using only first frames & 91.64 & 59.74 \\
When using subsampling of $s = 8$ & 82.88 & 90.21 \\
\midrule
\textit{Still} real images & 342.5 & 166.8 \\
\bottomrule
\end{tabular}
}
\end{table}

%% file: tables/inception-score.tex
\begin{table}[]
\caption{Inception Score~\cite{TGAN} on UCF101 $256^2$ (note that the underlying C3D model resizes the $256^2$ videos into $112^2$ resolution under the hood, eliminating high-quality details).}
\label{table:inception-score}
\centering
\begin{tabular}{lc}
\toprule
Method & Inception Score~\cite{TGAN} \\
\midrule
MoCoGAN~\cite{MoCoGAN} & 10.09$\pm$0.30 \\
MoCoGAN+SG2 \ours & 15.26$\pm$0.95 \\
VideoGPT~\cite{VideoGPT} & 12.61$\pm$0.33 \\
MoCoGAN-HD~\cite{MoCoGAN-HD} & 23.39$\pm$1.48 \\
DIGAN~\cite{DIGAN} & 23.16$\pm$1.13 \\
\modelname\ \ours & 23.94$\pm$0.73 \\
\midrule
Real videos & 97.23$\pm$0.38 \\
\bottomrule
\end{tabular}
\end{table}

%% file: tables/fid-scores.tex
\begin{table}[]
\caption{FVD$_{16}$, FID and training costs of modern video generators on FaceForensics $256^2$. Training cost is measured in terms of GPU-days.}
\label{table:fid-scores}
\centering
\resizebox{1.0\linewidth}{!}{
\begin{tabular}{lccc}
\toprule
Method & FVD$_{16}$ & FID & Training cost \\
\midrule
MoCoGAN~\cite{MoCoGAN} & 124.7 & 23.97 & 5 \\
MoCoGAN+SG2 \ours & 55.62 & 10.82 & 8 \\
VideoGPT~\cite{VideoGPT} & 185.9 & 22.7 & 32 \\
MoCoGAN-HD~\cite{MoCoGAN-HD} & 111.8 & 7.12 & 16.5 \\
DIGAN~\cite{DIGAN} & 62.5 & 19.1 & 16 \\
\modelname\ \ours & 47.41 & 9.445 & 8 \\
\midrule
StyleGAN2~\cite{StyleGAN2-ADA} & N/A & 8.42 & 7.72 \\
\bottomrule
\end{tabular}
}
\end{table}

%% file: appendix/failed-experiments.tex
\section{Failed experiments}\label{ap:failed-experiments}

In this section, we provide a list of ideas, which we tried to make work, but they didn't work either because the idea itself is not good, or because we didn't put enough experimental effort into investigating it.

\textbf{Hierarchical motion codes}.
We tried having several layers of motion codes. Each layer has its own distance between the codes. In this way, high-level codes should capture high-level motion and bottom-level codes should represent short local motion patterns. This didn't improve the scores and didn't provide any disentanglement of motion information. We believe that the motion should be represented differently (similar to FOMM~\cite{FOMM}), rather than with motion codes, because they make it difficult for $\G$ to make them temporily coherent.

\textbf{Maximizing entropy of motion codes to alleviate motion collapse}.
As an additional tool to alleviate motion collapse, we tried to maximize entropy of wave parameters of our motion codes.
The generator solved the task of maximizing the entropy well, but it didn't affect the motion collapse: it managed to save some coordination dimensions of $\bm v_t$ specifically to synchronize motions.

\textbf{Prorgressive growing of frequences in positional embeddings}.
We tried starting with low-frequencies first and progressively open new and new ones during the training.
It is a popular strategy for training implicit neural representations on reconstruction tasks (e.g., \cite{Nerfies, SAPE}), but in our case we found the following problem with it.
The generator learned to use low frequencies for representing high-frequency motion and didn't learn to utilize high frequencies for this task when they became available.
That's why high-frequency motion patterns (like blinking or speaking) were unnaturally slow.

\textbf{Continuous LSTM with EMA states}.
Our motion codes use sine/cosine activations, which makes them suffer from periodic artifacts (those artifacts are mitigated by our parametrization, but still present sometimes).
We tried to use LSTM, but with \emph{exponential moving average} on top of its hidden states to smoothen out motion representations temporally.
However, (likely due to the lack of experimental effort which we invested into this direction), the resulted motion representations were either too smooth or too sharp (depending on the EMA window size), which resulted in unnatural motions.

\textbf{Concatenating spatial coordinates}.
INR-GAN~\cite{INR-GAN} uses spatial positional embeddings and shows that they provide better geometric prior to the model.
We tried to use them as well in our experiments, but they didn't provide any improvement neither in qualitatively, nor quantitatively, but made the training slightly slower (by ${\approx}$\%10) due to the increased channel dimensionalities.

\textbf{Feature differences in $\D$}.
Another experiment direction which we tried is computing differences between activations of next/previous frames in a video and concatenating this information back to the activations tensor.
The intuition was to provide $\D$ information with some sort of ``latent'' optical flow information.
However, it made $\D$ too powerful (its loss became smaller than usual) and it started to outpace $\G$ too much, which decreased the final scores.

\textbf{Predicting $\delta^x$ instead of conditioning in $\D$}.
There are two ways to utilize the time information in $\D$: as a conditioning signal or as a learning signal.
For the latter one, we tried to predict the time distances between frames by training an additional head to predict the class (we treated the problem as classification instead of regression since there is a very limited amount of time distances between frames which $\D$ sees during its training).
However, it noticeably decreased the scores.

\textbf{Conditioning on video length}.
For \emph{unconditional} UCF-101, it might be very important for $\G$ to know the video length in advance.
Because some classes might contain very short clips (like, jumping), while others are very long, and it might be useful for $\G$ to know in advance which video it will need to generate (since we sample frames at random time locations during training).
However, utilizing this conditioning didn't influence the scores.

%% file: appendix/data.tex
\section{Datasets details}\label{ap:data}

\subsection{Datasets details}

We provide the dataset statistics in Fig~\ref{fig:data-video-lens} and their comparison in Table~\ref{table:datasets-info}.
Note, that for MEAD, we use only its front camera shots (originally, it releases shots from several camera positions).

\input{figures/data-video-lens}
\input{tables/datasets-info}

\subsection{Rainbow Jelly}
We noticed that modern video synthesis datasets are either too simple or too difficult in terms of content and motion, and there are no datasets ``in-between''.
That's why we introduce RainbowJelly: a dataset of ``floating'' jellyfish.
It is constructed from an 8-hour-long movie in 4K resolution and 30 FPS from the Hoccori Japan youtube video channel.
It contains simple content but complex hierarchical motions and this makes it a challenging but approachable test-bed for evaluating modern video generators.

For our RainbowJelly benchmark, we used the following film: \href{https://www.youtube.com/watch?v=P8Bit37hlsQ}{https://www.youtube.com/watch?v=P8Bit37hlsQ}.
We cannot release this dataset due to the copyright restrictions, but we released a full script which processes it (see the provided source code).
To construct a benchmark, we sliced it into 1686 chunks of 512 frames each, starting with the 150-th frame (to remove the loading screen), center-cropped and resized into $256^2$ resolution.
This benchmark is advantageous compared to the existing ones in the following way:
\begin{enumerate}
    \item It contains complex hierarchical motions:
    \begin{itemize}
        \item a jellyfish flowing in a particular direction (low-frequency global motion);
        \item a jellyfish pushing water with its arms (medium-frequency motion)
        \item small perturbations of jellyfish's body and tentacles (high-frequency local motion).
    \end{itemize}
    \item It is a very high-quality dataset (4K resolution).
    \item It is simple in terms of content, which makes the benchmark more focused on motions.
    \item It contains long videos.
\end{enumerate}

%% file: figures/data-video-lens.tex
\newcommand{\dataVideoLensSubfigureScale}{0.7}
\newcommand{\dataVideoLensSubfigureVspace}{-0.3cm}

\begin{figure}
\centering
\begin{subfigure}[b]{\dataVideoLensSubfigureScale\linewidth}
    \centering
    \includegraphics[width=\textwidth]{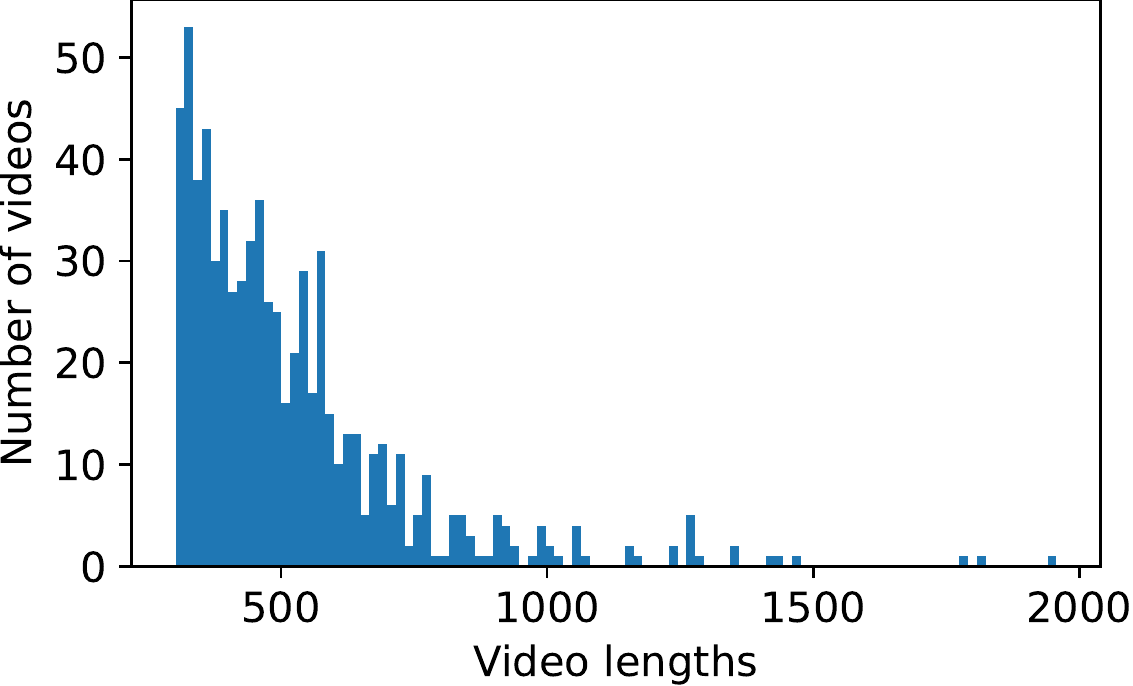}
    \caption{FaceForensics~\cite{FaceForensics_dataset}.}
\end{subfigure}

\vspace{\dataVideoLensSubfigureVspace}
\hfill

\begin{subfigure}[b]{\dataVideoLensSubfigureScale\linewidth}
    \centering
    \includegraphics[width=\textwidth]{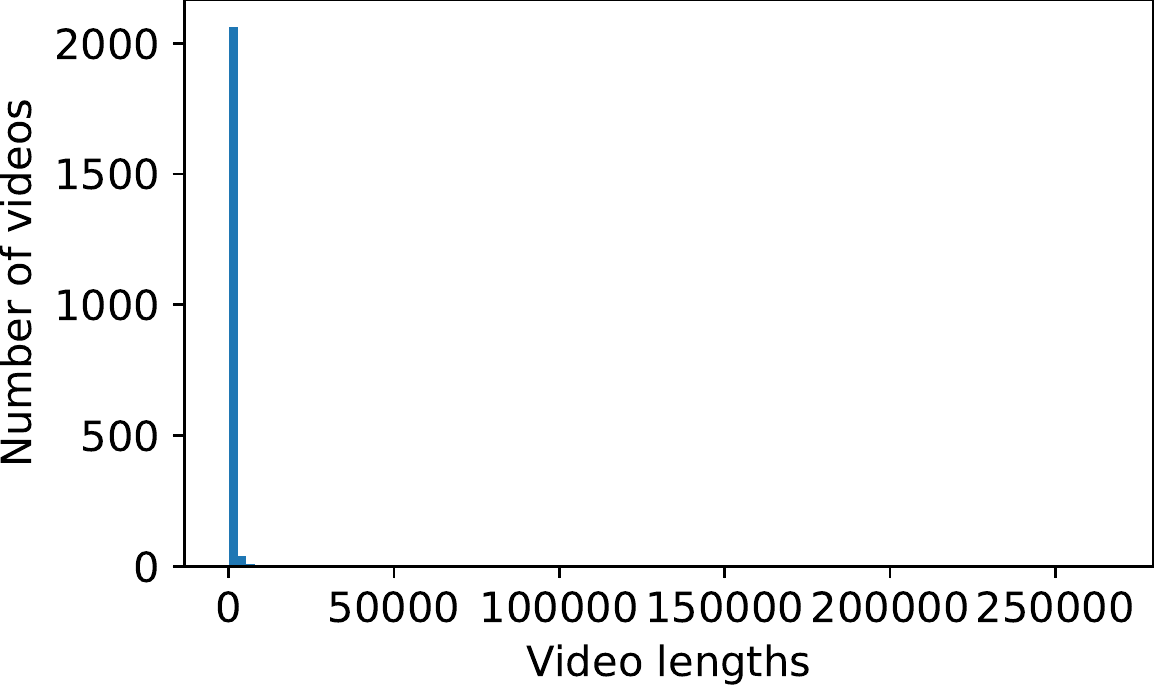}
    \caption{SkyTimelapse~\cite{SkyTimelapse_dataset}.}
\end{subfigure}

\vspace{\dataVideoLensSubfigureVspace}
\hfill

\begin{subfigure}[b]{\dataVideoLensSubfigureScale\linewidth}
    \centering
    \includegraphics[width=\textwidth]{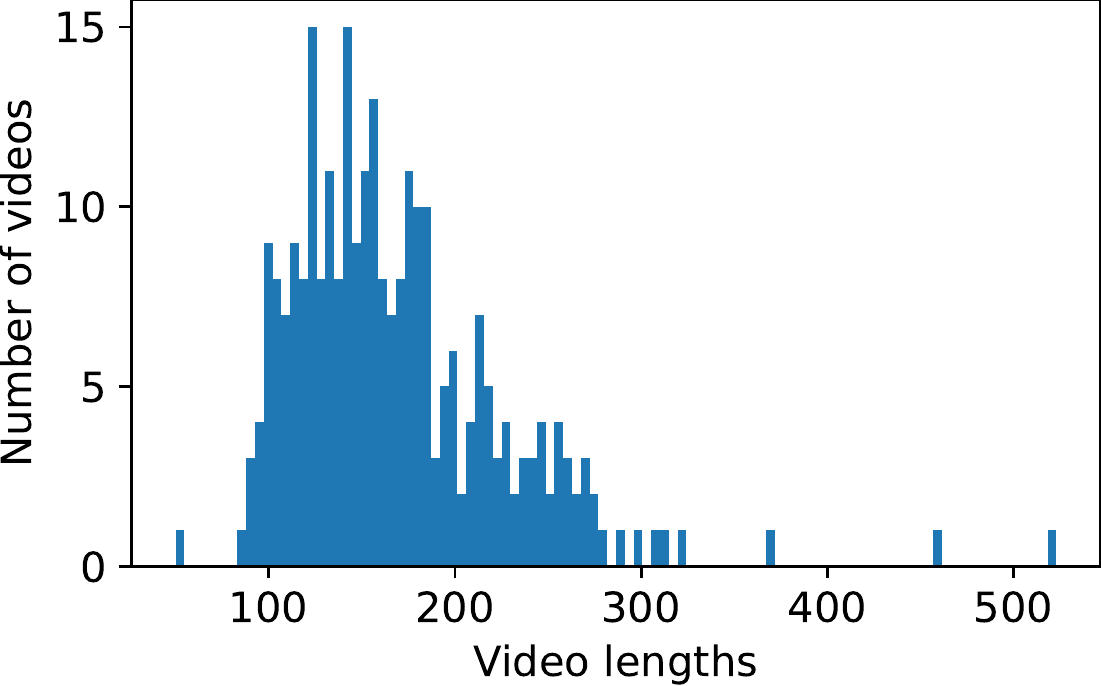}
    \caption{UCF-101~\cite{UCF101_dataset}.}
\end{subfigure}

\vspace{\dataVideoLensSubfigureVspace}
\hfill

\begin{subfigure}[b]{\dataVideoLensSubfigureScale\linewidth}
    \centering
    \includegraphics[width=\textwidth]{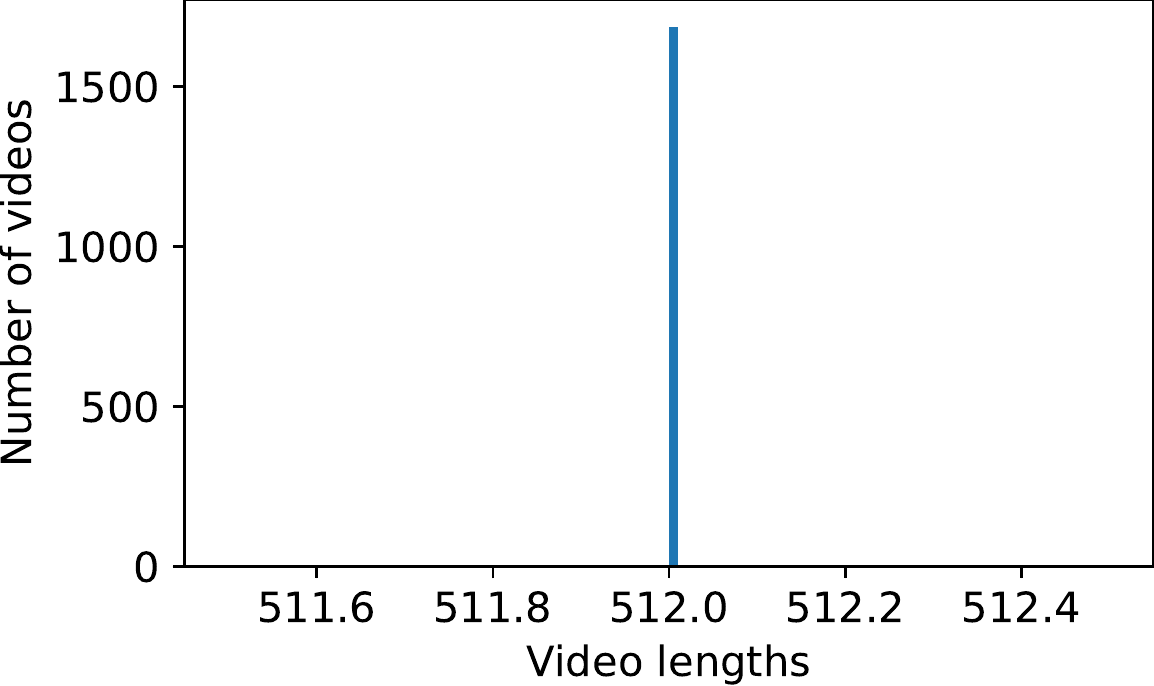}
    \caption{RainbowJelly~\cite{UCF101_dataset}.}
\end{subfigure}

\vspace{\dataVideoLensSubfigureVspace}
\hfill

\begin{subfigure}[b]{\dataVideoLensSubfigureScale\linewidth}
    \centering
    \includegraphics[width=\textwidth]{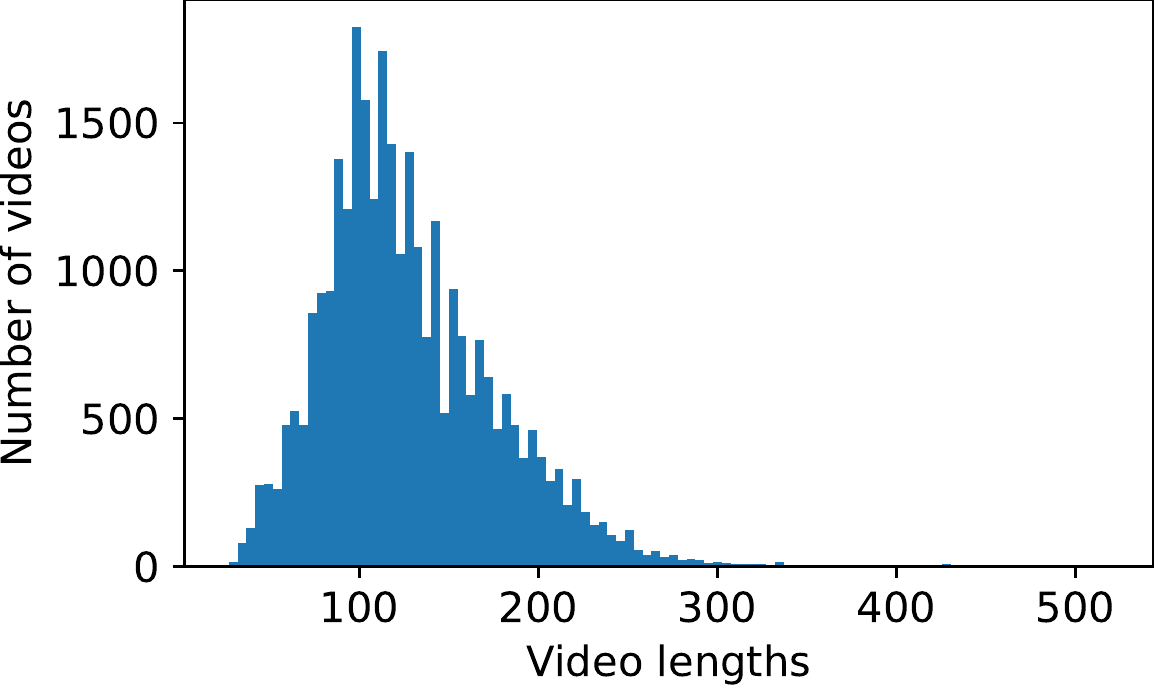}
    \caption{MEAD~\cite{UCF101_dataset}.}
\end{subfigure}
\caption{Distribution of video lengths (in terms of numbers of frames) for different datasets. Note that RainbowJelly and MEAD~\cite{MEAD_dataset} are 30 FPS, while the rest are 25 FPS datasets. Note that SkyTimelapse contains several very long videos which might bias the distribution if not treated properly.}
\label{fig:data-video-lens}
\vspace{-0.5cm}
\end{figure}

%% file: tables/datasets-info.tex
\begin{table}
\caption{Additional datasets information in terms of total lengths (in the total number of hours), average video length (in seconds), frame rate and the amount of speakers (for FaceForensics and MEAD).}
\label{table:datasets-info}
\centering
\resizebox{1.0\linewidth}{!}{
\begin{tabular}{lcccc}
\toprule
Dataset & \#hours & avg len & FPS & \#speakers \\ 
\midrule
FaceForensics~\cite{FaceForensics_dataset} & 4.04 & 20.7s & 25 & $704$ \\
SkyTimelapse~\cite{SkyTimelapse_dataset} & 12.99 & 22.1s & 25 & N/A \\
UCF-101~\cite{UCF101_dataset} & 0.51 & 6.8s & 25 & N/A \\
RainbowJelly & 7.99 & 17.1s & 30 & N/A \\
MEAD~\cite{MEAD_dataset} & 36.11 & 4.3s & 30 & $48$ \\
\bottomrule
\end{tabular}
}
\end{table}

%% file: appendix/assumptions.tex
\section{Implicit assumptions of sparse training}\label{ap:assumptions}

In this section, we elaborate on our simple theoretical exposition from \S\ref{sec:method:sampling}

Consider that we want to fit a probabilistic model $q_\theta(\bm x)$ to the real data distribution $\bm x \sim p(\bm x) = p(x_1, ..., x_n)$.
For simplicity, we will be considering a discrete finite case, i.e. $n < \infty$, but note that videos, while continuous and infinite in theory, are still discretized and have a time limit to fit on a computer in practice.
For fitting the distribution, we use \textit{$k$-sparse training}, i.e. picking only $k$ random coordinates from each sample $\bm x \sim p(\bm x)$ during the optimization process.
In other words, introducing $k$-sparse sampling reformulates the problem from
\begin{equation}\label{eq:normal-opt-problem}
d(p(\bm x), q_\theta(\bm x)) \longrightarrow \min_\theta
\end{equation}
into
\begin{equation}\label{eq:sparse-opt-problem}
\sum_{I \in \mathcal{I}^k} d(p(\bm x_I), q_\theta(\bm x_I)) \longrightarrow \min_\theta,
\end{equation}
where $d(\cdot, \cdot)$ is a problem-specific distance function between probability distributions, $\mathcal{I}^k$ is a collection of all possible sets $I = \{i_1, ..., i_k\}$ of unique indices $i_j \in \{1, 2, ..., n\}$ and $\bm x_I$ denotes a sub-vector $(x_{i_1}, ..., x_{i_k})$ of $\bm x$.
This means, that instead of bridging together full distributions we choose to bridge all their possible marginals of length $k$ instead.
When solving Eq.~\eqref{eq:sparse-opt-problem} will help us to obtain the full joint distribution $p(\bm x)$?
To investigate this question, we develop the following simple statement.

Let's denote by $\mathcal{J}_{<i}^k$ a collection of sets $J_i$ of up to $k$ indices s.t. $\forall J_i \in \mathcal{J}_{<i}^k$ we have $j < i$ for all $j \in J_i$.

Using the chain rule, we can represent $p(\bm x)$ as:
\begin{equation}
p(\bm x) = \prod_{i=1}^n p(x_i | \bm x_{<i}),
\end{equation}
where $\bm x_{<i}$ denotes the sequence $(x_1, ..., x_{i-1})$.
Now, if we know that for each $i$, there exists $J_i = \{j_1, ..., j_{k-1}\}$ with $j_\ell < i$ s.t.:

\begin{equation}\label{eq:short-history}
p(x_i | \bm x_{<i}) = p(x_i | \bm x_{J_i}),
\end{equation}
then $p(\bm x)$ is obviously simplified to:
\begin{equation}
p(\bm x) = \prod_{i=1}^n p(x_i | \bm x_{<i}) = \prod_{i=1}^n p(x_i | \bm x_{J_i})
\end{equation}

Does this tell anything useful?
Surprisingly, yes.
It says that if $p(\bm x)$ is simple enough that instead of using the whole history $\bm x_{<i}$ to model $p(x_i | \bm x_{<i})$ it's enough to use only some set ``representative moments'' $J_i$ (unique for each $i$) with the size $|J_i| < k$, then $k$-sparse training is a viable alternative.
After fitting $q_\theta(\bm x)$ via $k$-sparse training, we will be able to obtain $p(\bm x)$ using Eq~\eqref{eq:short-history} \emph{even though} $q_\theta(\bm x)\not\equiv p(\bm x)$!
Note, that one can obtain a conditional distributional $p(x_i | \bm x_I)$ from the marginal one $p(x_i, \bm x_I)$ for some set of indicies $I = \{ i_1, ..., i_{\ell-1} \}$ via:
\begin{equation}\label{eq:conditional-to-marginal}
p(x_i | \bm x_I) = \frac{p(x_i, \bm x_I)}{p(\bm x_I)} = \frac{p(x_i, \bm x_I)}{ \int_{x_i} p(x_i, \bm x_I) d x_i }.
\end{equation}

But we would also like to have the ``reverse'' dependency, i.e. knowing that if we can approximate the distribution via a set of marginals, then this distribution is not too difficult.
For this claim, we will need to consider marginals not of an arbitrary form $p(\bm x_S)$, but of the form $p(x_i , J_i)$, and we would need exactly $n$ of those.
The reverse implication is the following.
\textit{If $p(\bm x)$ can be represented as a product of $n$ conditionals $p(i | J_i)$, then for each $i$ there exists $J_i \in \mathcal{J}^k_{<i}$ s.t. $p(x_i | \bm x_i) = p(x_i | J_i)$.}
This statement, just like the previous one, looks obvious.
But oddly, requires more than a single sentence to prove.
First, we are given that:
\begin{equation}
p(\bm x) = \prod_{i=1}^n p(x_i | \bm x_{<i}) = \prod_{i=1}^n p(x_i | \bm x_{J_i}),
\end{equation}
but unfortunately, we cannot directly claim that each term in the product $\prod_{i=1}^n p(x_i | \bm x_{<i})$ equals to its corresponding one in the product $\prod_{i=1}^n p(x_i | \bm x_{J_i})$.
For this, we first need to show that for each $m$ we have:
\begin{equation}
p(\bm x_{\leq m}) = \prod_{i=1}^m p(x_i | \bm x_{J_i})
\end{equation}
It can be seen from the fact, that:
\begin{equation}
\begin{split}
p(\bm x_{\leq m}) &=
\int_{\bm x_{>m}} p(\bm x) d \bm x_{>m} \\
&=
\int_{\bm x_{>m}} \prod_{i=1}^n p(x_i | \bm x_{J_i}) d \bm x_{>m} \\
&=
\prod_{i=1}^m p(x_i | \bm x_{J_i}) \cdot \int_{\bm x_{>m}} \prod_{i=m+1}^n p(x_i | \bm x_{J_i}) d \bm x_{>m} \\
&=
\prod_{i=1}^m p(x_i | \bm x_{J_i}) \cdot 1 \\
&=
\prod_{i=1}^m p(x_i | \bm x_{J_i})
\end{split}
\end{equation}

This allows to cancel terms in the chain rule one by one, starting from the end, leading to the desired equality:
\begin{equation}
p(x_i | \bm x_{<i}) = p(x_i | \bm x_{J_i})
\end{equation}

Does this reverse claim tells us anything useful?
Surprisingly again, yes.
It implies that if we managed to fit $p(\bm x)$ by using $k$-sparse training, then this distribution is not sophisticated.

Merging the above two statements together, we see that \textit{$p(\bm x)$ can be represented as a product of $n$ conditionals $p(x_i | \bm x_{J_i})$ for $i = 1, ..., n$ if and only if for all $i \leqslant n$ there exists $J_i \in \mathcal{J}_{<i}^{k-1}$ s.t. $p(x_i | \bm x_{<i}) \equiv p(x_i | \bm x_{J_i})$}.

What does this statement tell for video synthesis?
Any video synthesis algorithm utilizes $k$-sparse training to learn its underlying model, but in contrast to prior work, we use very small values of $k$.
This means, that we fit our model $q_\theta(\bm x)$ to model any $k$-marginals of $p(\bm x)$ (considering that we pick frames uniformly at random) instead of the full one $p(\bm x)$.
And using the above statement, such a setup implies the assumption of Eq~\eqref{eq:short-history}.
This equation says that one can know everything about $x_i$ by just observing previous frames $J_i$.
In other words, $x_i$ must be predictable from $\bm x_{J_i}$.
Moreover, it is easy to show that our statement can generalize to include several $J_i^{(1)}, ..., J_i^{(\ell)}$ for $i$-th frame, i.e. there might exist several explainable sets of frames.

%% file: appendix/additional-samples.tex
\section{Additional samples}\label{ap:additional-samples}
For the ease of visualization, we provide additional samples of the model via a web page: \href{\websiteurl}{\websiteurl}.

%% file: appendix/digan.tex
\section{Comparison to DIGAN}\label{ap:digan}

Our model shares a lot of similarities to DIGAN~\cite{DIGAN} and in this section we highlight those similarities and differences.

\subsection{Major similarities}
\textbf{Sparse training}. DIGAN also utilizes very sparse training (only 2 frames per video).
But in our case, we additionally explore the optimal number of frames per video $k$ (see \S\ref{sec:method:sampling}).

\textbf{Continuous-time generator}. DIGAN also builds a generator, which is continuous in time.
But our generator does not lose the quality at infinitely large lengths.

\textbf{Dropping \texttt{conv3d} blocks}. DIGAN also drops \texttt{conv3d} blocks in their discriminator.
But in contrast to us, they still have 2 discriminators.

\subsection{Major differences}
\textbf{Motion representation}. DIGAN uses only a single global motion code, which makes it \emph{theoretically} impossible to generate infinite videos: at some point it will start repeating itself (due to the usage of sine/cosine-based positional embeddings). In our case, we use an infinite sequence of motion codes, which are being temporally interpolated, computed wave parameters from and transformed into motion codes. DIGAN mixes temporal and spatial information together into the same positional embedding, which creates the following problem: even when time changes, the spatial location, perceived by the model, \textit{also changes}. This creates a ``head-flying-away'' effect (see the samples). In our case, we keep these two information sources decomposed from one another.

\textbf{Generator's backbone}. DIGAN is built on top of INR-GAN~\cite{INR-GAN}, while our work uses StyleGAN2. This allows DIGAN to inherit INR-GAN's benefits from being spatially continuous, but at the expense of being less stable and being slower to train (due to the lack of mixed precision and increased channel dimensionalities from concatenating positional embeddings).

\textbf{Discriminator structure}. DIGAN uses \emph{two} discriminators: the first one operates on image-level and is equivalent to StyleGAN2's one, while the other one operates on ``video'' level and takes frames $\bm x_{t_1}, \bm x_{t_2}$ and the time differences between them $\Delta = t_2 - t_1$, concatenates them all together into a 7-channel input image (tiling the time difference scalar) and passes into a model with StyleGAN2 discriminator's backbone. In our case, we use concatenate the frames features and apply the conditioning via the projection discriminator~\cite{ProjectionDiscriminator} strategy.

\textbf{Sampling procedure}. We use $k=3$ samples per video, while DIGAN uses $k=2$. Also, we sample frames uniformly randomly, while DIGAN selects $t_1 \sim \text{Beta}(2, 1)$ and $t_1 \sim \text{Beta}(1, 2)$ (in this way, DIGAN sometimes have $t_1 > t_2$). Apart from that, they use $T=16$.

Apart from those major distinctions, there are lot of small implementation differences.
We refer an interested reader to the released codebases for them:
\begin{itemize}
    \item \modelname: \href{\codeurl}{\codeurl}
    \item DIGAN: \href{https://openreview.net/forum?id=Czsdv-S4-w9}{https://openreview.net/forum?id=Czsdv-S4-w9}
\end{itemize}

\subsection{A note on the computational cost}

INR-GAN demonstrated that it has higher throughput than StyleGAN2 in terms of images/second~\cite{INR-GAN}.
But the authors compare to the original StyleGAN2 implementation and not to the one from StyleGAN2-ADA repo, which is \emph{much} better optimized.
Also, they use caching of positional embeddings which is only possible at test-time and has great influence on its computational performance.
In this way, we found that that StyleGAN2 is ${\approx}2$ times faster to train and is \emph{less} consuming in terms of GPU memory than INR-GAN.

DIGAN is based on top of INR-GAN and that's why suffers from the issues described above.
We trained it for a week on $\times 4$ v100 NVidia GPUs and observed that it stopped improving after ${\approx}5$ days of training.
This is equivalent to ${\approx}20k$ real frames seen by the discriminator (while MoCoGAN+SG2 and \modelname\ reach ${\approx}25k$ in just 2 days for the same resolution in the same environment).
For the time of the submitting the main paper, there was no information about the training cost.
However, the authors updated their manuscript for the time of submitting the supplementary and specify the training cost of 8 GPU-days \emph{$128^2$ resolution}, which is consistent with our experiments (considering that we have twice as larger resolution).